%% file: main.tex
\documentclass{article}

\let\OriginalAddContentsLine\addcontentsline
\usepackage{iclr2027_conference,times}
\let\addcontentsline\OriginalAddContentsLine

\usepackage[T1]{fontenc}
\usepackage[utf8]{inputenc}
\usepackage{microtype}
\usepackage{amsmath,amssymb,mathtools}
\usepackage{graphicx}
\usepackage{booktabs,tabularx,array,multirow}
\usepackage{float}
\usepackage{subcaption}
\usepackage[svgnames]{xcolor}
\usepackage{xspace,comment}
\usepackage{tikz}
\usetikzlibrary{arrows.meta,positioning,fit}
\usepackage[normalem]{ulem}
\usepackage{wrapfig}
\usepackage{enumitem}
\usepackage[most]{tcolorbox}
\usepackage{hyperref}
\hypersetup{hidelinks}
\usepackage[all]{hypcap}
\usepackage{etoolbox}
\makeatletter
\patchcmd{\@maketitle}
  {\vskip 0.3in minus 0.1in}
  {\vskip 0.12in}
  {}
  {\PackageWarning{layout}{Author spacing patch failed}}
\makeatother

\input{case_preamble}
\input{macro}


\title{Fly-by-Code: Embodied Coding Agents for Aerial Manipulation with Active Visual and Physical Feedback}

\author{%
\textbf{Jaewoo Lee$^{1*}$, Jeongyeon Seo$^{1*}$, Sihyun Cho$^{1*}$, Gyeongrak Choe$^{1}$, Yutong Wang$^{2}$,}\\
\textbf{Bavin Saravanan$^{2}$, Jia-Bin Huang$^{3}$, Furong Huang$^{3,4}$, Sebastian Scherer$^{2}$, Guanya Shi$^{2}$,}\\
\textbf{H.~Jin Kim$^{1}$, Seungjae Lee$^{3\dagger}$, Dongjae Lee$^{5\dagger}$}\\[6pt]
\textnormal{%
$^{1}$Seoul National University
\hspace{1.5em}
$^{2}$Carnegie Mellon University
}\\[2pt]
\textnormal{%
$^{3}$University of Maryland, College Park
\hspace{1.5em}
$^{4}$All Purpose AI
\hspace{1.5em}
$^{5}$Kyung Hee University
}
}
\date{}

\iclrfinalcopy

\begin{document}

\addtocontents{toc}{\protect\setcounter{tocdepth}{-1}}

\maketitle
\lhead{Preprint}

\begingroup
\renewcommand{\thefootnote}{}
\footnotetext{$^{*}$Equal contribution; order determined by coin flip. $^{\dagger}$Corresponding authors.}
\endgroup

\begin{figure}[!htbp]
    \centering
    \includegraphics[width=\linewidth]{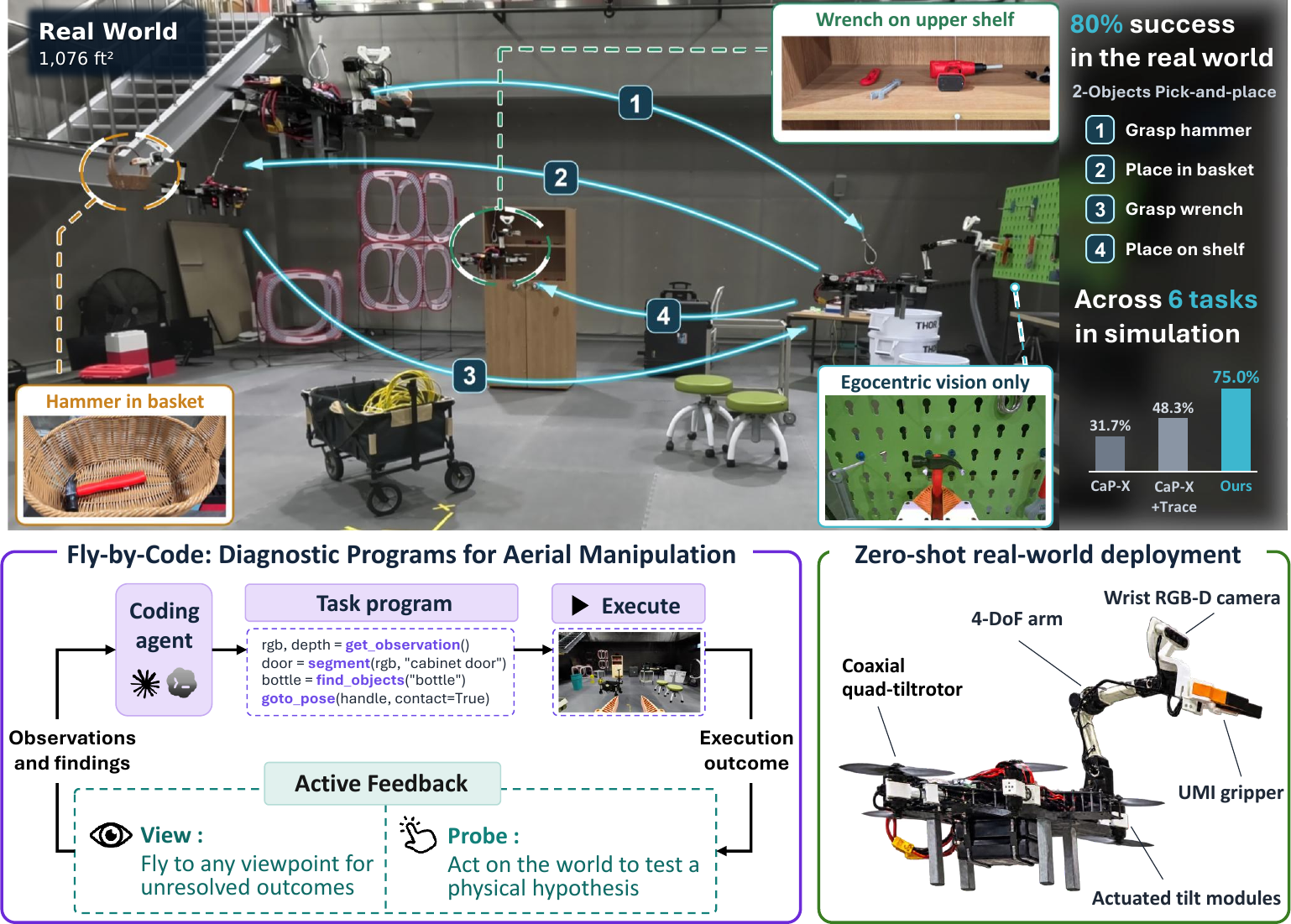}
    \caption{\algoname uses \textbf{Active Feedback} (AF) to address missing outcome observations and unresolved failure causes arising from partial observability and limited interaction robustness.
    Rather than relying only on feedback passively collected during task execution, Fly-by-Code generates an executable diagnostic program to acquire the evidence needed for outcome assessment and program revision.
    The program either navigates to an informative viewpoint (\textbf{View}) to inspect an unobserved outcome or performs a targeted physical interaction (\textbf{Probe}) to distinguish between possible failure causes.
    Using shared robot APIs, task programs refined in simulation run on a physical aerial manipulator without source-code modification and complete multi-stage tasks such as the illustrated four-step tool-manipulation sequence.}
    \label{k2_real_world_thumbnail}
\end{figure}

\input{sections/abstract}
\input{sections/introduction}
\input{sections/related_work}
\input{sections/method}
\input{sections/experiments}
\input{sections/realworld}
\input{sections/conclusion}
\input{sections/limitations}
\input{sections/acknowledgement}

\clearpage
\bibliographystyle{iclr2027_conference}
\bibliography{references}

\clearpage

\appendix
\addtocontents{toc}{\protect\setcounter{tocdepth}{2}}
\renewcommand{\contentsname}{\Large Appendix}
\setcounter{tocdepth}{2}
\tableofcontents
\clearpage
\input{sections/appendix}

\end{document}

%% file: case_preamble.tex
\usepackage{graphicx}
\tcbuselibrary{most} 
\usepackage{listings}

\definecolor{caseframe}{HTML}{9AABBB}
\definecolor{caseback}{HTML}{F7F9FC}
\definecolor{casetitle}{HTML}{DCE6EF}
\definecolor{casetitletext}{HTML}{263F57}

\definecolor{codebase}{HTML}{30343B}
\definecolor{codekeyword}{HTML}{435A78}
\definecolor{codestring}{HTML}{47766C}
\definecolor{codecomment}{HTML}{858B94}
\definecolor{codefunction}{HTML}{594F72}

\newtcolorbox{caseevidence}[1]{
    enhanced,
    breakable,
    colback=caseback,
    colframe=caseframe,
    colbacktitle=casetitle,
    coltitle=casetitletext,
    title={#1},
    fonttitle=\normalfont\normalsize\bfseries,
    fontupper=\normalfont\normalsize,
    boxrule=0.45pt,
    arc=1.5mm,
    left=7pt,
    right=7pt,
    top=5pt,
    bottom=5pt,
    toptitle=3pt,
    bottomtitle=3pt,
    before skip=7pt,
    after skip=7pt
}

\lstdefinestyle{casecode}{
    basicstyle=\normalfont\normalsize\color{codebase},
    columns=fullflexible,
    keepspaces=true,
    showstringspaces=false,
    breaklines=true,
    breakatwhitespace=true,
    breakindent=12pt,
    aboveskip=2pt,
    belowskip=2pt,
    xleftmargin=0pt,
    xrightmargin=0pt,
    keywordstyle=\color{codekeyword},
    stringstyle=\color{codestring},
    commentstyle=\color{codecomment},
    emph={
        look,
        get_observation,
        goto_pose,
        open_gripper,
        release_and_hold,
        report,
        asarray,
        array,
        pos,
        go
    },
    emphstyle=\color{codefunction}
}

\providecommand{\casepagebreak}{}

%% file: macro.tex
\newcommand{\algoname}{Fly-by-Code\xspace}
\newcommand{\AF}{\textsc{AF}\xspace}
\newcommand{\AFlong}{Active Feedback\xspace}

\definecolor{instructionframe}{HTML}{A4B4C0}
\definecolor{instructiontitle}{HTML}{E8EEF3}
\definecolor{instructionbody}{HTML}{F8FAFC}
\definecolor{instructiontext}{HTML}{304858}

\newcounter{instructiontask}

\newtcolorbox{taskinstructionbox}[1]{
    enhanced,
    title={#1},
    colback=instructionbody,
    colframe=instructionframe,
    colbacktitle=instructiontitle,
    coltitle=instructiontext,
    fonttitle=\small\bfseries,
    fontupper=\small,
    boxrule=0.4pt,
    arc=1.5mm,
    boxsep=0pt,
    left=9pt,
    right=9pt,
    top=6pt,
    bottom=6pt,
    toptitle=5pt,
    bottomtitle=5pt,
    before skip=0pt,
    after skip=0pt,
    before upper={\setlength{\parindent}{0pt}}
}

\newcommand{\taskentry}[3]{%
    \par\addvspace{12pt}%
    \noindent
    \begin{minipage}{\linewidth}
        \setlength{\parindent}{0pt}%
        \setlength{\parskip}{0pt}%
        \refstepcounter{instructiontask}%
        \begin{taskinstructionbox}{Task \theinstructiontask: #1}
            #2
        \end{taskinstructionbox}
        \par\vspace{5pt}%
        \noindent\textbf{Setup.} #3\par
    \end{minipage}%
    \par
}

%% file: sections/abstract.tex
\begin{abstract}
Embodied coding agents construct robot policies from natural-language instructions by generating executable programs and refining them from execution feedback. This refinement assumes that execution feedback reveals whether the task succeeded and, after a failure, provides sufficient evidence to guide program revision.
Aerial manipulation violates this assumption in two ways. Payload constraints restrict onboard camera coverage and leave task-critical outcomes outside the robot’s view (partial observability), while strong dynamic coupling and limited control margins make physical interactions fragile (limited interaction robustness) and leave failure causes unresolved. Re-examining the same feedback cannot recover observations that were never collected or separate failure causes that the available evidence does not distinguish. To acquire this missing evidence, we introduce Active Feedback (AF), an active “View and Probe” framework. When execution feedback leaves unresolved questions, the agent synthesizes a separate diagnostic program that runs in simulation to inspect unseen outcomes from informative viewpoints (View) or test hypotheses about failure causes through targeted physical interactions (Probe). 
Across six simulated aerial manipulation tasks with 60 trials per condition, AF reduces false completion declarations from 25 to 5 and raises final task success from 31.7\% to 75\% compared with terminal feedback under the same coding-agent configuration and task-execution limit. Using shared robot APIs, task programs refined with AF in simulation run on a physical aerial manipulator without task-program source-code modification and complete multi-stage manipulation tasks in the real world. These results show that executable diagnosis fills gaps in execution feedback for embodied coding agents, improving both completion assessment and program refinement in aerial manipulation.
\textbf{Project Page:}
{\urlstyle{same}\url{https://fly-by-code.github.io/}}

\vspace{2mm}

\end{abstract}

%% file: sections/introduction.tex
\section{Introduction}
\label{sec:intro}

Embodied coding agents provide a promising approach to task-level coordination by translating natural-language instructions into executable programs that combine general control flow and computation with calls to robot perception and action APIs~\citep{liang2023code, chen2024roboscript, fu2026capx}. In an iterative refinement loop, the agent executes a program, examines execution feedback, and revises it until it determines the task is complete. This refinement implicitly assumes that execution feedback provides sufficient evidence to assess the task outcome and, when execution fails, to identify how to revise the program.

This assumption breaks down under two challenging conditions: partial observability and limited interaction robustness. Partial observability can leave task-critical outcomes outside the robot's observations, while failures arising from physical interaction limitations can be difficult to distinguish from perception or manipulation errors. We refer to these two gaps in execution feedback as \emph{missing outcome observations} and \emph{unresolved failure causes}, respectively. Re-examining the same feedback cannot supply observations that were never collected or resolve failure causes that the available evidence does not distinguish. Effective refinement therefore requires the agent to identify what additional evidence is needed and how to acquire it.

Aerial manipulation amplifies both limitations of execution feedback. An aerial robot equipped with a manipulator must physically interact with the environment while maintaining flight~\citep{ollero2022past}. Spatially extended tasks can exceed the coverage of a fixed external camera, leaving the agent dependent on egocentric onboard sensing whose viewpoint changes with robot motion. At the same time, limited control margin and disturbances from manipulator motion and interaction wrenches can produce failures~\citep{yadav2025modular, hwang2025autonomous} that are difficult to distinguish from perception or manipulation errors. For example, an empty gripper after a placement attempt does not reveal whether the object reached the target, while a failed door-opening attempt may not reveal whether the grasp failed, the commanded motion was inappropriate, or the platform lacked sufficient physical capability.

Resolving these gaps requires new evidence: choosing a viewpoint that reveals an unseen outcome or performing a targeted physical interaction that tests a specific failure hypothesis. Motivated by this observation, we introduce \textbf{\AFlong} (\AF), an explicit \emph{post-execution diagnostic stage}. Rather than immediately revising the task program based on existing execution feedback, AF first lets the agent generate and execute a separate diagnostic program to gather evidence for unresolved questions. Using the same robot APIs as the task program, the diagnostic program can move to informative viewpoints to acquire missing outcome observations (\textbf{View}) or perform targeted physical interactions in simulation to test hypotheses about unresolved failure causes related to interaction behavior or embodiment constraints (\textbf{Probe}). The resulting observations and measurements are returned to the same agent, providing additional evidence for both completion assessment and subsequent program revision.

Existing coding-agent systems obtain evidence through multimodal execution feedback~\citep{fu2026capx}, per-primitive traces~\citep{lu2026aspire}, task-specific verification procedures~\citep{xiao2026enpire}, or information-gathering actions embedded within task execution~\citep{fu2026capx}. AF instead synthesizes an attempt-specific diagnostic procedure after execution, separately from task evaluation. AF uses the same prompt across tasks, while the agent generates a diagnostic program for each attempt to investigate unresolved questions. By decoupling diagnosis from task progress, AF gives the agent freedom to select actions for their information value rather than their immediate contribution to task completion. Task outcomes are evaluated before diagnosis, and the environment resets before the next task attempt, so diagnostic actions contribute evidence without earning task credit.

Across six aerial manipulation tasks in two simulated environments, AF achieves 75\% final task success, outperforming terminal (31.7\%) and execution-trace (48.3\%) feedback under the same agent configurations and execution limits. These results show that additional observations and interaction evidence improve program synthesis beyond existing execution-feedback baselines. We analyze task progress across refinement iterations, evaluate AF across models and interfaces, and examine how diagnostics inform program revisions.

Finally, we investigate the real-world transferability of our approach by deploying the synthesized programs on a physical aerial manipulator. Because simulation and hardware share task-level APIs, these programs run on our custom hardware without source-code changes, even though the underlying sensing and control implementations differ. The resulting programs repeatedly complete two multi-stage tasks, 2-Objects Pick-and-place and Cabinet, despite differences between the simulated and physical scenes and variations in initial robot position and heading across trials (Figure~\ref{k2_real_world_thumbnail}).

In summary, our work makes the following key contributions:
\begin{itemize}
    \item We characterize two limitations of execution feedback for embodied coding agents: \emph{missing outcome observations} and \emph{unresolved failure causes}, which hinder completion assessment and program revision.
    \item We introduce \textbf{Active Feedback} (AF), a post-execution diagnostic stage within a simulation-based program-refinement loop. At this stage, the agent synthesizes an attempt-specific executable program to acquire evidence through informative viewpoints that supply missing outcome observations (\textbf{View}) and targeted physical interactions that test hypotheses about unresolved failure causes (\textbf{Probe}).
    \item We evaluate AF in aerial manipulation. Across six simulated tasks, AF improves final task success to 75\%, compared with 31.7\% for terminal feedback and 48.3\% for execution-trace feedback, while also improving completion assessment. We further demonstrate repeated physical execution of simulation-synthesized task programs through shared robot APIs without task-program source-code modification. To the best of our knowledge, this is the first study of \emph{embodied coding agents for aerial manipulation}.
\end{itemize}

%% file: sections/related_work.tex
\section{Related Work}
\label{sec:related}

\paragraph{Robot program synthesis and feedback-driven refinement.}
Code-as-policy and program-synthesis approaches represent robot behavior as executable programs that compose perception and control APIs \citep{liang2023code,singh2023progprompt,chen2024roboscript, mu2024robocodex,oelerich2025languageguided,mower2026rosllm}. Recent work extends program generation through iterative refinement: CaP-X uses textual and visual execution feedback, ASPIRE leverages multimodal execution traces with reusable repairs and program search, ARCHITECT incorporates execution traces and human corrections into reusable skills, and ENPIRE constructs task-specific reset, safety, and verification interfaces \citep{fu2026capx,lu2026aspire,chen2026few,xiao2026enpire}. Related methods explain failures from execution experience, generate recovery behaviors, or acquire missing physical information \citep{liu2023reflect,schlesinger2024roborepair,lin2026physcap}. Active Feedback (AF) instead addresses cases where existing execution evidence is insufficient, synthesizing an attempt-specific diagnostic program to actively acquire visual or physical evidence for completion assessment and subsequent program refinement.

\paragraph{Active and interactive perception.}
Active perception selects sensing actions to acquire task-relevant information \citep{bajcsy1988active,bajcsy2018revisiting}, while interactive perception uses physical interaction to reveal information that is difficult to obtain from passive observation alone \citep{bohg2017interactive}. Recent robotic manipulation systems extend these ideas through task-driven viewpoint selection and coordinated perception--manipulation under occlusion \citep{weng2024interactive,wang2025observe,xiong2025vision, liu2026activevla,liu2026sapave}. Language-model-based agents can likewise select epistemic actions to acquire multimodal information under partial observability \citep{zhao2023chat}. Whereas prior methods primarily acquire information to support ongoing perception or task execution, AF synthesizes a separate diagnostic program after each attempt to resolve uncertainties for completion assessment and program refinement.

\paragraph{Learning-based and language-guided aerial manipulation.}
Aerial manipulation brings together challenges in flight,
perception, and physical interaction, with research spanning
platform design, contact-aware control, whole-body motion planning,
and perception-driven manipulation~\citep{ollero2022past, bodie2021active, zhao2023dragon, guo2024flying, deng2025whole,
ubellacker2024highspeed, bauer2025opensource, lee2026autonomous}. Learning-based approaches learn manipulation policies from
demonstrations or adapt pretrained policies to aerial platforms
\citep{he2025flying, gupta2026umi, tucker2026pimakefly},
while we focus on actively acquiring post-execution evidence
to assess and revise policies expressed as executable programs. Among language-guided approaches, AERMANI-VLM constrains high-level decisions to predefined
flight-safe skills to improve execution reliability \citep{mishra2025aermani}, while AeroGrab actively changes viewpoints to improve target observability under occlusion \citep{singh2026aerograb}.

%% file: sections/method.tex
\section{Active Feedback for Program Synthesis in Aerial Manipulation}
\label{sec:method}
\begin{figure}[t]
    \centering
    \includegraphics[width=\linewidth]{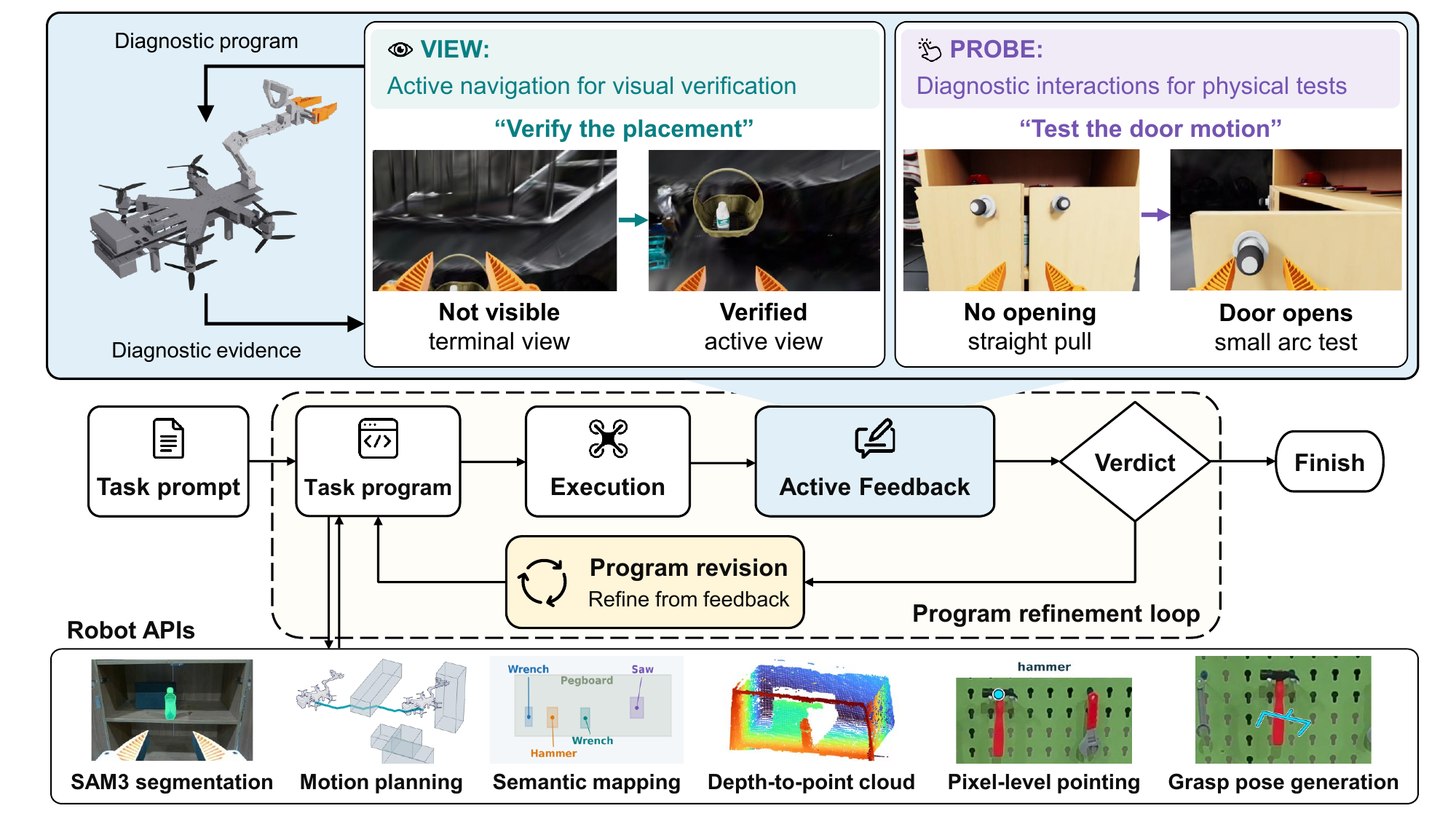}
    \vskip -0.06in
    \caption{Overview of \algoname. \textbf{Bottom:} The program refinement loop. The coding agent translates a task prompt into an executable task program using shared robot APIs. Following execution, it enters the Active Feedback (AF) stage to acquire diagnostic evidence, iteratively refining the program until it decides to terminate the revision process. \textbf{Top:} During AF, the agent synthesizes a diagnostic program to resolve execution uncertainties. \textbf{View} actively navigates to informative viewpoints (e.g., inspecting a basket from above to verify placement). \textbf{Probe} conducts diagnostic physical tests (e.g., testing door articulation with an arc motion after a straight pull fails).}
    \vskip -0.2in
    \label{fig:overview}
\end{figure}

Active Feedback (AF) distinguishes \algoname from existing embodied coding agents that rely solely on \emph{passive} execution feedback for completion assessment and program refinement.
Such feedback is limited to evidence collected during task execution.
AF inserts an \emph{active} diagnostic stage before the decision to finish or revise the task program: the agent writes a separate program that runs in simulation to acquire additional evidence for unresolved questions (Figure~\ref{fig:overview}).

\subsection{Motivation: Limitations of Passive Feedback}

Passive execution feedback includes console logs and terminal observations. In aerial manipulation, this feedback alone has two fundamental limitations. First, partial observability leaves \emph{missing outcome observations}. Because an egocentric camera tightly couples its viewpoint to the robot's flight trajectory, critical end-states frequently fall outside the terminal field of view. Second, limited interaction robustness leaves \emph{unresolved failure causes}. A static terminal image may not distinguish whether an interaction failed due to erroneous contact geometry or violated kinematic constraints. To fill these gaps, \algoname supplements passive observation logs with actively acquired evidence.

\subsection{Diagnostic Program Synthesis in AF}
Following a task attempt, the agent reviews the execution logs, terminal observations, and semantic-map summary within AF to identify unverified task outcomes and unresolved failure causes. It then writes a diagnostic program to investigate these unresolved questions. This program uses the same shared robot APIs (Appendix~\ref{app:apis}, Table~\ref{tab:robot-apis}) as the task program but serves a fundamentally different purpose: acquiring information rather than advancing the task. Because the environment records the task score before this stage, diagnostic interactions do not affect the evaluation. By synthesizing diagnostic code after task execution, the agent can target uncertainties specific to the current attempt rather than relying on pre-programmed verification steps (Appendix~\ref{app:k2_case}).

\subsection{Diagnostic Actions: View and Probe}

\begin{figure}[t]
    \centering
    \includegraphics[width=\linewidth]{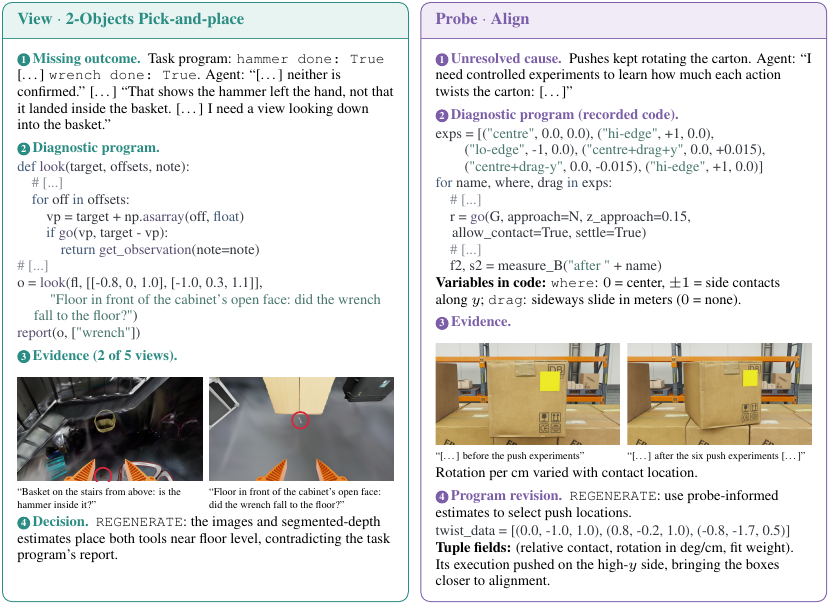}
    \caption{
        Recorded \textbf{View} and \textbf{Probe} rounds, abridged from Appendix~\ref{app:cases}.
        \textbf{Left:} The task program reported both tools placed.
        The diagnostic program acquired views of the destinations and nearby floor, which place both tools near floor level, and the agent rejected the reported success.
        \textbf{Right:} After pushes kept rotating a carton, the agent compared contact responses across six push tests.
        The next task program converts the measurements into a push-response model for contact selection.
        Quotations, code, and image notes are recorded agent outputs, with omissions marked by [\ldots] or \texttt{\# [...]}.
        Red circles were added by the authors.
    }
    \label{fig:view-probe-examples}
\end{figure}

Diagnostic programs acquire two kinds of evidence, one for each gap in execution feedback: \textbf{View} acquires missing outcome observations, and \textbf{Probe} tests hypotheses about unresolved failure causes.
Neither is a fixed routine or a dedicated API.
The agent writes both from the same robot APIs as the task program, decides what to inspect or test in the current attempt, and may combine them in one diagnostic program.
Figure~\ref{fig:view-probe-examples} traces a recorded example of each, from the agent's unresolved question to the returned evidence and the resulting decision.

\paragraph{View: Active Observation of Task Outcomes.}
Task outcomes may be spatially distributed and remain outside the terminal camera view. A View program flies the wrist camera to viewpoints selected to answer specific questions about the outcome, using semantic-map queries when helpful to locate relevant objects.
Each selected image is returned with an agent-written note stating the question it is meant to answer, and measurements such as segmented-depth object positions are returned through console output.
In 2-Objects Pick-and-place, for example, views of the destinations and the surrounding floor showed that both tools had fallen despite the task program's success report, and the agent selected \texttt{REGENERATE} (Figure~\ref{fig:view-probe-examples}, left; Appendix~\ref{app:k2_case}).

\paragraph{Probe: Diagnostic Physical Interaction.}
Some physical uncertainties cannot be resolved through observation alone, such as how an object responds to contact, how an articulated door moves, or whether a load exceeds the platform's lift capacity.
A Probe program performs physical interactions chosen for the information they provide and measures the responses.
In Align, for example, the agent compared pushes at different contact points and used the measured rotations to select the contact position in the next task program (Figure~\ref{fig:view-probe-examples}, right; Appendix~\ref{app:align_case}).

\noindent Diagnostic programs start from the state left by the task program, so a probe can examine the stage at which execution failed without replaying earlier steps.
Because the task score is already recorded and the environment is reset before the next attempt, probes may also include actions that would be undesirable during task execution, such as attempting a lift that may exceed the platform's capacity.

\subsection{Closing the Feedback Loop}
The feedback loop concludes by integrating diagnostic findings into completion assessment and subsequent program revision. The diagnostic program returns selected observations, agent-written inspection notes, and console measurements to the agent, which evaluates them alongside the original execution logs to declare task completion (\texttt{FINISH}) or synthesize a revised task program (\texttt{REGENERATE}). Before the next task attempt, the environment resets its physical state, while the diagnostic information persists within the agent's context to inform the revised task program.

%% file: sections/experiments.tex
\section{Experiments}
\label{sec:experiments}
We evaluate Active Feedback (AF) through three questions: (Q1) Does AF improve task success and completion assessment over passive execution feedback? (Q2) How does diagnostic evidence inform program refinement? (Q3) How does AF perform across models and agent interfaces? Section~\ref{sec:sim-to-real} then evaluates real-world deployment of the simulation-synthesized programs.

\subsection{Experimental Setup}

\paragraph{Simulation Environments and Robot.} We evaluate \algoname in six tasks extending AM-Bench \citep{wang2026ambench}, implemented in Isaac Lab~\citep{nvidia2025isaaclabgpuacceleratedsimulation}. The simulated agent controls an omnidirectional aerial manipulator (Appendix \ref{app:system}). The tasks range from spatial object transport (1-Object and 2-Objects Pick-and-place, Sort) to contact-rich interactions involving articulated structures (Cabinet), unknown payload constraints (Payload), and contact-dependent pushing (Align). A shared execution backend manages collision-aware whole-body motion planning and tracks the generated trajectories under simulated flight dynamics (Appendix~\ref{app:apis}).
\begin{figure}[t]
    \centering
    \includegraphics[width=\linewidth]{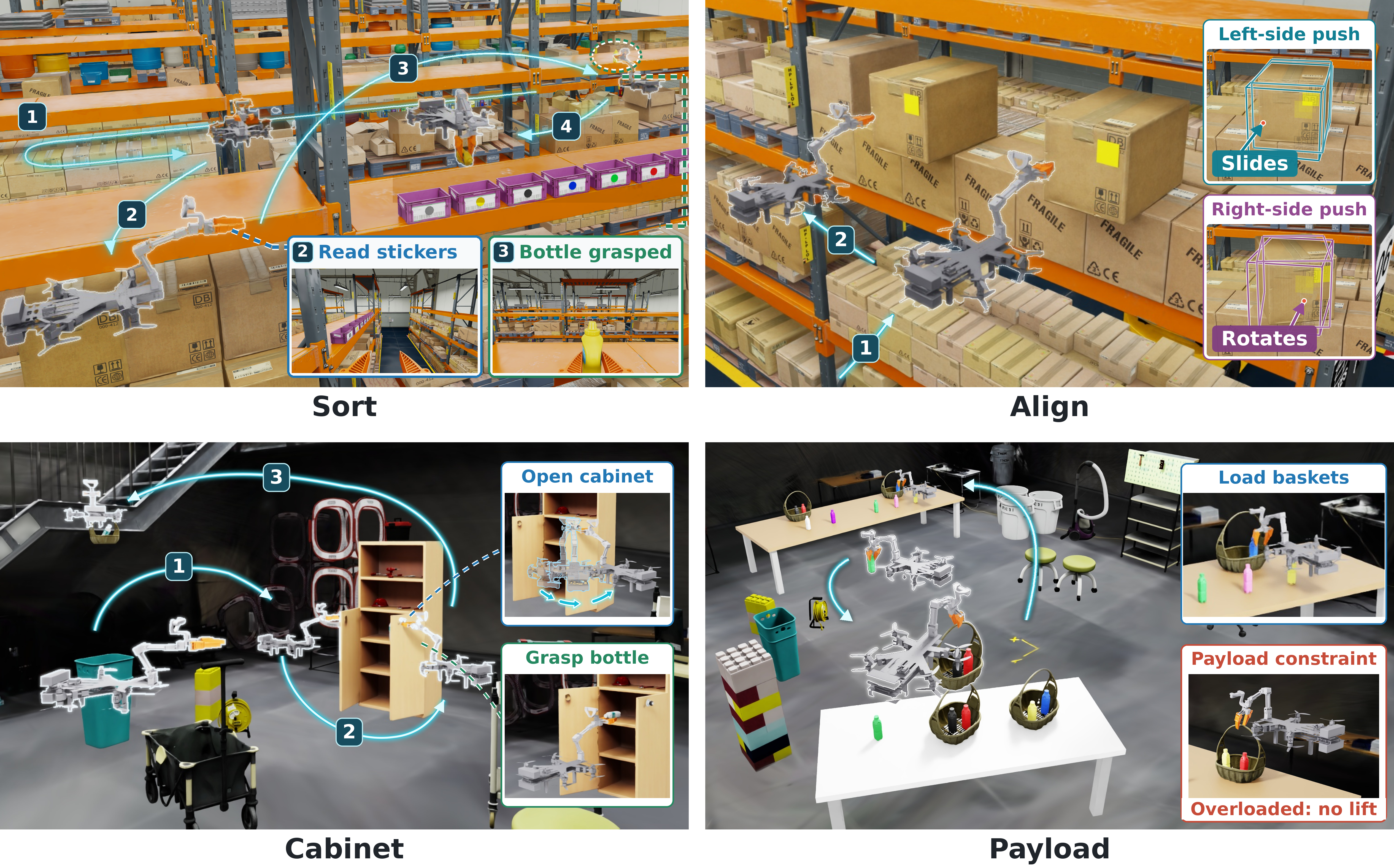}
    \caption{Representative aerial manipulation tasks. \textbf{Sort:} Identify the target bottle and place it into the color-matched bin. \textbf{Align:} Push protruding boxes into alignment with the rack.
    Because the centers of mass are offset, varying contact points induce distinct translational or rotational motions.
    \textbf{Cabinet:} Open the cabinet doors, retrieve the bottle inside, and deliver it to a hanging basket. \textbf{Payload:} Transport filled bottles and baskets with minimal hall crossings under an undisclosed payload limit. The Sort and Align tasks operate in the warehouse simulation, whereas Cabinet and Payload use the laboratory digital twin. Overlaid robot poses and arrows illustrate sequential execution stages and motion trajectories.}
    \label{k2_real_world_results}
\end{figure}

\paragraph{Baselines and Agent Configuration.}

We build on CaP-X~\citep{fu2026capx}, adapting its program-generation and execution loop to aerial manipulation. All conditions share the same perception, control, and semantic-map APIs and a fixed skill library adapted from CaP-X. We use direct RGB feedback without CaP-Agent0's visual-differencing module or parallel ensemble. Unlike CaP-X's continuous execution protocol, we execute each revised program from the initial environment state, and it must complete the full task. Within this common loop, we compare AF against two baselines: CaP-X, which provides execution logs and terminal RGB observations, and CaP-X+Trace, which additionally provides structured primitive-call records and execution keyframes. The agent receives these traces before deciding whether to finish or regenerate the task program. CaP-X+Trace adapts the trace-based feedback used in ASPIRE~\citep{lu2026aspire} but does not incorporate its skill library or evolutionary program search. The primary evaluation uses Claude Opus~5~\citep{anthropic2026opus5} through persistent sessions with effort set to \texttt{high}. Each trial permits one initial task-program execution and up to ten revisions.

\subsection{Task Success and Completion Assessment}

\begin{figure}[t]
  \centering
  \includegraphics[width=\linewidth]{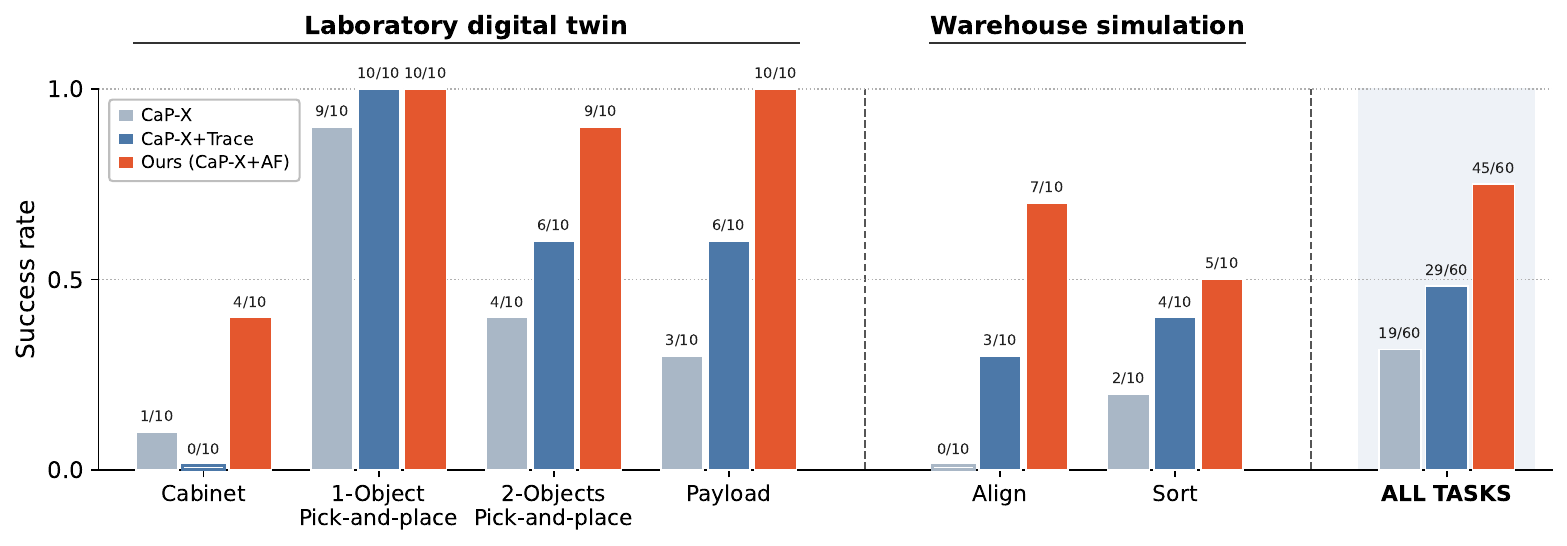}
  \vskip -0.1in
  \caption{Final task success across six tasks. AF achieves the highest success rate overall and matches or exceeds both CaP-X and CaP-X+Trace on every task.
  Each trial permits one initial execution and up to ten program revisions. Labels above the bars indicate successful trials out of ten total trials per task.}
  \vskip -0.1in
  \label{fig:task-success}
\end{figure}

\paragraph{Overall Task Performance.} Across six aerial manipulation tasks, AF achieves an overall success rate of 75\%, significantly outperforming CaP-X (31.7\%) and CaP-X+Trace (48.3\%) (Figure \ref{fig:task-success}). This advantage becomes increasingly pronounced in tasks requiring extended physical sequences. For example, in 2-Objects Pick-and-place, AF reaches 90\% success compared to 40\% and 60\% for the baselines. Similarly, AF substantially improves performance on tasks governed by hidden physical constraints, such as Payload (100\% vs. 30\% and 60\%) and Cabinet (40\% vs. 10\% and 0\%). Beyond final success rates, AF also accelerates average task progress, showing that active diagnostic findings enable more effective program refinement than passive traces (Figure~\ref{fig:task-progress} in Appendix~\ref{app:task-progress}).

\paragraph{Reliable Completion Assessment.} Crucially, AF drastically reduces false positive completion assessments. Across 60 trials, AF yields only five false positives, achieving completion-assessment precision of 90.0\% (Table \ref{tab:finish-outcomes}). In contrast, CaP-X and CaP-X+Trace produce 25 (43.2\% precision) and 19 (60.4\% precision) false positives, respectively. While baseline agents frequently declare success based on unverified execution logs—such as assuming an open gripper guarantees a successful placement—AF actively acquires visual and physical evidence to verify spatially distributed outcomes before making a final judgment.

\input{tables/false_positive}

\subsection{How Does Active Feedback Support Program Refinement?}

\begin{figure}[t]
    \centering
    \includegraphics[width=\linewidth]{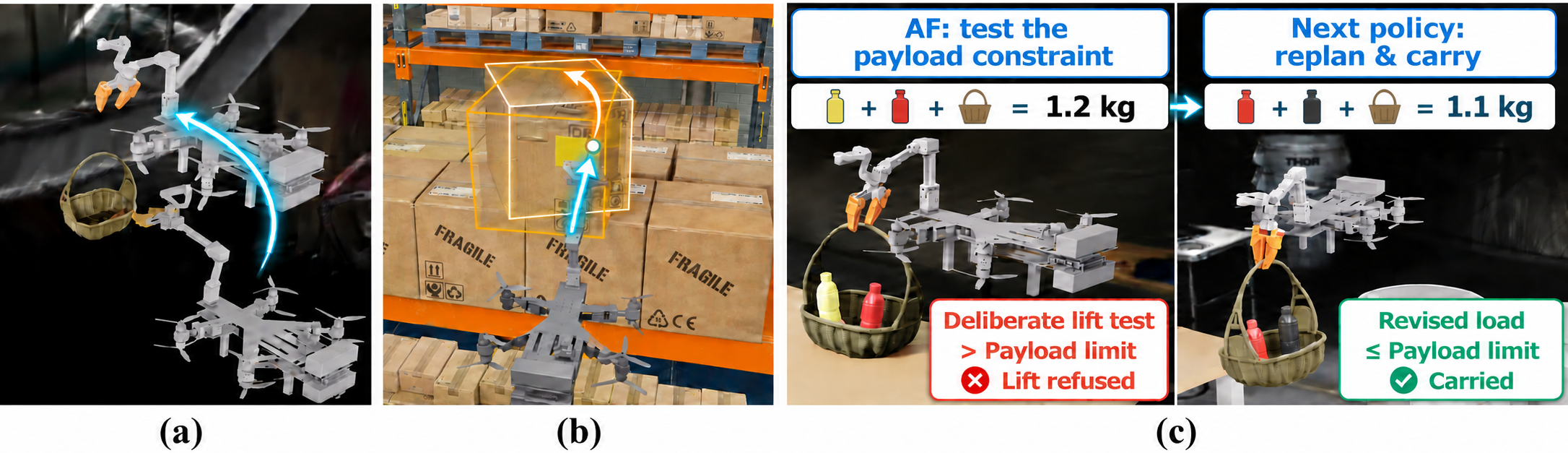}
    \vskip -0.1in
    \caption{Examples of Active Feedback (AF) for completion assessment and program refinement. \textbf{(a) View:} The robot moves above the basket to inspect the placement outcome. \textbf{(b) Probe:} A targeted push reveals the box's response to physical contact. \textbf{(c) Probe:} A diagnostic lift test of a 1.2\,kg basket load is refused, indicating that the proposed load is infeasible (left). The next task program revises the object grouping and successfully carries a 1.1\,kg load (right).}
    \vskip -0.1in
    \label{fig:payload-case}
\end{figure}

To analyze how AF evidence informs the agent's decisions, we examine examples from three tasks (Figure~\ref{fig:payload-case}). Building on the diagnostic actions in Section~3.3, these examples show how View evidence informs completion assessment and how Probe evidence guides program revision at the control or planning level.

\textbf{View: Verifying Task Outcomes.}
In 2-Objects Pick-and-place, an empty gripper after release does not confirm successful placement. In a recorded AF attempt, the agent moved the wrist camera to inspect the basket, shelves, and surrounding floor to check the tools' actual locations (Figure~\ref{fig:payload-case}(a); Figure~\ref{fig:view-probe-examples}, left; Appendix~\ref{app:k2_case}). Although the task program reported success without visual confirmation that the tools had reached their intended destinations, the AF observations revealed unsuccessful placements and led the agent to select \texttt{REGENERATE}. Across ten trials per condition on this task, AF produced one false completion declaration, compared with five for CaP-X and three for CaP-X+Trace (Table~\ref{tab:finish-outcomes}). 

\textbf{Probe: Refining Control and Planning.} The agent used Probe evidence to infer interaction models and planning constraints that guided subsequent program revision. At the control level, the Align agent compared pushes at different contact points and used the measured responses to initialize a push-response model for contact selection in the next program (Figure~\ref{fig:payload-case}(b); Figure~\ref{fig:view-probe-examples}, right). At the planning level, the Payload agent tested whether it could lift a proposed 1.2\,kg basket load. The refused lift led the next program to regroup the bottles under a conservative 1.1\,kg planning limit, and both revised basket loads were lifted successfully (Figure~\ref{fig:payload-case}(c), Appendix~\ref{app:payload_case}). In both cases, targeted tests informed specific revisions within the full-task refinement loop. 

\subsection{Generality Across Models and Interfaces}

\begin{figure}[t]
    \centering
    \includegraphics[width=\linewidth]
        {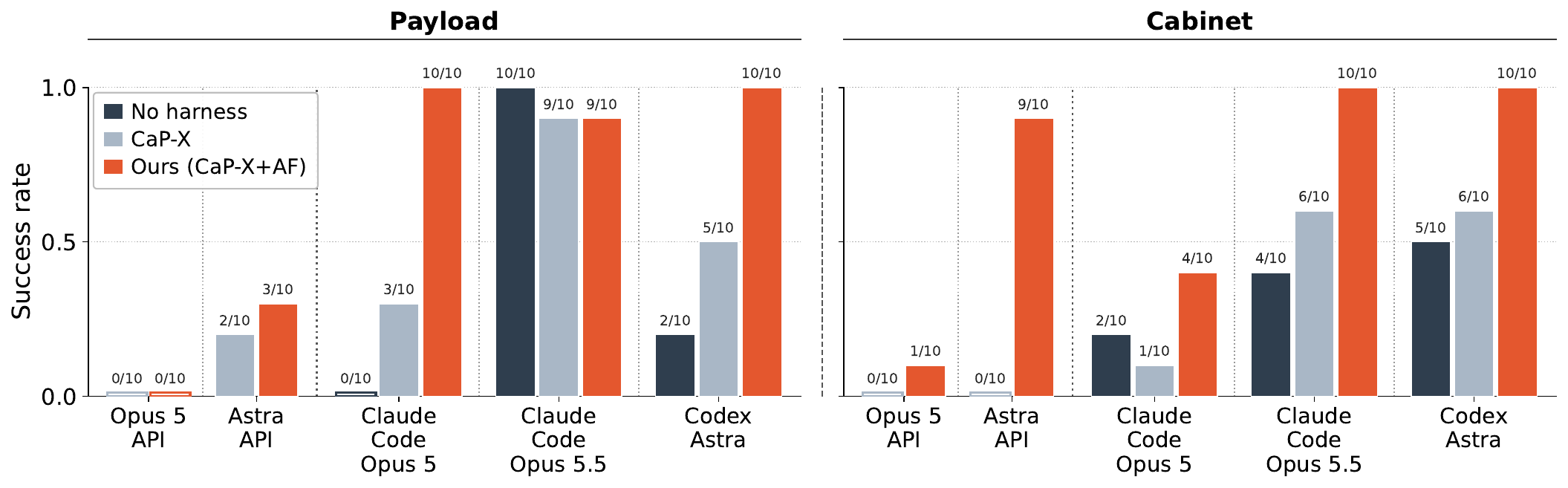}
    \vskip -0.1in
    \caption{
        Final task success across models and agent interfaces on Payload
    and Cabinet. Labels above the bars indicate successful trials /
    total trials, with ten trials per task and condition.
    }
    \vskip -0.1in
    \label{fig:api-ablation-success}
\end{figure}

\paragraph{Consistent Performance Gains.}
Active Feedback (AF) improves task success across multiple foundation models and agent interfaces (Figure~\ref{fig:api-ablation-success})~\citep{anthropic2026opus5, anthropic2026opus55, openai2026astra}. Across all evaluated task--backbone model pairs, AF matches or exceeds the success rate of the passive CaP-X baseline. On the Cabinet task, AF drastically raises success from 0\% to 90\% using the Astra API, and from 60\% to 100\% with both Claude Code Opus 5.5 and Codex Astra. Similarly, on the Payload task, AF elevates success from 30\% to 100\% with Claude Code Opus 5, and from 50\% to 100\% with Codex Astra. These results show that AF’s benefits extend across multiple model–interface configurations, although their magnitude varies.

\paragraph{Necessity of Structured Diagnostics.}
To isolate the impact of the structured diagnostic stage in AF, we evaluate a ``No harness'' baseline that receives a superset of execution evidence---including full raw videos and console logs---and freely inspects it without predefined feedback packaging. Despite this unrestricted access to information from all sensors, the No harness condition struggles on the Cabinet task, achieving 20\% to 50\% across coding-agent configurations and reaching at most half the success rate of AF. On the Payload task, No harness achieves 0\% with Claude Code Opus 5 and 20\% with Codex Astra, whereas AF achieves 100\% with both. While unstructured data access occasionally benefits specific model-task combinations (e.g., Payload with Claude Code Opus 5.5), the substantial overall performance gap demonstrates that synthesizing targeted diagnostic programs resolves execution uncertainties more effectively than unstructured log analysis.

%% file: tables/false_positive.tex
\begin{table}[h]
  \centering
  \caption{Completion assessment performance. Each triplet reports correct completion declarations (TP), incorrect completion declarations (FP), and trials reaching the step limit without a declaration (ND), with 10 trials per task and 60 overall. Precision is TP/(TP+FP). P\&P denotes Pick-and-place.}
  \label{tab:finish-outcomes}
  \resizebox{0.8\textwidth}{!}{%
  \renewcommand{\arraystretch}{1.15}
  \begin{tabular}{@{}l cccccc | cr@{}}
    \toprule
    & \multicolumn{6}{c|}{Task Outcomes (TP / FP / ND)} & \multicolumn{2}{c}{Overall} \\
    \cmidrule(lr){2-7} \cmidrule(l){8-9}
    Method & Cabinet & 1-Obj P\&P & 2-Obj P\&P & Payload & Align & Sort & TP/FP/ND & Precision \\
    \midrule
    CaP-X
    & 1/\textbf{1}/8 & 9/1/0 & 4/5/1
    & 3/6/1 & 0/10/0 & 2/2/6 & 19/25/16 & 43.2\% \\
    CaP-X+Trace
    & 0/3/7 & \textbf{10}/\textbf{0}/0 & 6/3/1
    & 6/3/1 & 3/7/0 & 4/3/3 & 29/19/12 & 60.4\% \\
    \midrule
    \textbf{AF (Ours)}
    & \textbf{4}/\textbf{1}/5 & \textbf{10}/\textbf{0}/0 & \textbf{9}/\textbf{1}/0
    & \textbf{10}/\textbf{0}/0 & \textbf{7}/\textbf{3}/0 & \textbf{5}/\textbf{0}/5 & \textbf{45}/\textbf{5}/10 & \textbf{90.0\%} \\
    \bottomrule
  \end{tabular}
  }
\end{table}

%% file: sections/realworld.tex
\subsection{Sim-to-Real Transfer and Physical Deployment}
\label{sec:sim-to-real}

To demonstrate \algoname's practical robustness, we deploy task programs synthesized with AF in simulation on a physical aerial manipulator. Physical aerial manipulation involves dynamic disturbances and visual conditions that differ from those in simulation. However, through a unified hardware-software interface, the physical robot executes the exact simulation-synthesized Python programs without modifying the source code. Rather than replaying simulated trajectories, the deployed task programs use real-world observations through the shared robot APIs to guide execution.

\input{tables/realworld}

\paragraph{Autonomous Long-Horizon Execution.}
We evaluate this zero-shot physical transfer on the 2-Objects Pick-and-place and the Cabinet tasks. The unmodified programs successfully coordinate sustained perception, flight, and contact-rich interactions in the real world. As detailed in Table \ref{tab:hardware-task-outcomes}, \algoname achieves a full-task success rate of 80\% (4/5 trials) for 2-Objects Pick-and-place and 60\% (3/5 trials) for the Cabinet task. Notably, these executions require approximately 8 to 9 minutes of sustained autonomous operation. Repeated task completion despite differences between simulated and physical scene configurations and variations in initial robot pose supports the transfer of task programs refined with AF in simulation (Appendix~\ref{app:hardware_trials}, Tables~\ref{tab:sim-real-object-positions} and~\ref{tab:hardware-start-poses}).

%% file: tables/realworld.tex
\begin{table}[htbp]
    \centering
    \caption{
    Physical task outcomes. Denominators include all evaluated trials per task, even those not reaching the subtask.
    }
    \label{tab:hardware-task-outcomes}
    \small
    \begin{tabular}{llc}
        \toprule
        Task & Outcome & Successful trials \\
        \midrule
        2-Objects Pick-and-place & Grasp hammer & 5/5 \\
                            & Place hammer in basket & 4/5 \\
                            & Grasp wrench & 5/5 \\
                            & Place wrench on shelf & 5/5 \\
                            & \textbf{Full task} & \textbf{4/5} \\
        \midrule
        Cabinet & Open left door & 5/5 \\
                              & Open right door & 4/5 \\
                              & Grasp bottle & 4/5 \\
                              & Place bottle in basket & 3/5 \\
                              & \textbf{Full task} & \textbf{3/5} \\
        \bottomrule
    \end{tabular}%
\end{table}

%% file: sections/conclusion.tex
\section{Conclusion}

We introduce Active Feedback (AF), a post-execution diagnostic stage in which the agent synthesizes a separate program to acquire informative viewpoints and conduct targeted physical tests in simulation. The resulting evidence informs both completion judgments and subsequent program refinement. We study this approach in embodied coding agents for aerial manipulation, where long-horizon task execution requires reliable completion assessment and informative feedback for program revision. Across six simulated tasks, AF improves final task success and completion assessment compared with terminal and execution-trace feedback. Through shared robot APIs, task programs refined with AF in simulation achieve zero-shot sim-to-real transfer to a physical aerial manipulator without modifying task-program source code, repeatedly completing long-horizon tasks despite differences in scene configuration and variations in initial robot pose. More broadly, these results suggest that embodied coding agents benefit from actively acquiring the evidence they need, rather than relying only on feedback collected during task execution.

%% file: sections/limitations.tex
\section{Limitations}

While Active Feedback (AF) helps \algoname acquire missing outcome observations and test hypotheses about unresolved failure causes, this explicit diagnostic stage adds model-inference and robot-interaction overhead during diagnosis. Furthermore, although post-execution inspection actively acquires targeted evidence, its effectiveness remains bounded by onboard perception errors and unreachable viewpoints. Simply performing a diagnostic check does not guarantee a correct task-completion judgment if the underlying sensory data is flawed. In \algoname, diagnostic interactions occur in simulation, and the refined task programs achieve zero-shot sim-to-real transfer without modifying the task-program source code. However, this protocol does not provide an operational safety guarantee for unrestricted physical probing directly in the real world. Additionally, while the framework effectively handles spatial discrepancies across known tasks, our successful deployment relies heavily on underlying motion planning and flight controllers. We do not claim zero-shot generalization to entirely unseen physical environments or radically different embodiment morphologies.

%% file: sections/acknowledgement.tex
\section*{Acknowledgements}
We thank Sungjin Kim for extensive assistance with the experiments and Dr.~Youngseok Jang for insightful discussions and support throughout this project. \\ Guanya Shi holds concurrent appointments as an Assistant Professor at Carnegie Mellon University and as an Amazon Scholar. This paper describes work performed at Carnegie Mellon University and is not associated with Amazon. Furong Huang and Seungjae Lee are supported by DARPA HR001124S0029-AIQ-FP-019, National Science Foundation TRAILS Institute (2229885). Private support was provided by Open Philanthropy and Apple. The authors acknowledge the National Artificial Intelligence Research Resource (NAIRR) Pilot for contributing to this research result.

%% file: sections/appendix.tex
\section{Implementation Details}
\label{app:implementation}

\subsection{Program synthesis and active feedback}
\label{app:algorithm}

The coding agent initially receives a task instruction, an embodiment description, task-specific environment notes, robot API documentation, and an initial wrist-camera RGB image. Active Feedback (AF) diagnostic programs execute from the state left by the task program, without an intervening reset. Before each revised task program runs, the environment and semantic map are reset. Task outcomes are evaluated before AF, as described in Appendix~\ref{app:scoredefs}.

\paragraph{Evidence review.}
The AF prompt includes the executed task program, its console output, the terminal camera image, and the current semantic-map summary. It instructs the agent to assess the available evidence for each task requirement before generating diagnostic code:
\begin{quote}
\small
FIRST, REASON --- in plain text, BEFORE any code: take the task PART BY PART and say, for each part, whether what you already have settles it. If you are not sure, it is not settled.
[\ldots]
THEN ACTIVELY INSPECT what you could not confirm.
\end{quote}
If the existing evidence confirms every requirement, the agent may instead return \texttt{NO\_INSPECTION\_NEEDED} with a brief explanation, skipping diagnostic execution.

\paragraph{Diagnostic interaction.}
The AF prompt explicitly directs the agent to acquire informative viewpoints and permits information-seeking physical interaction in simulation:
\begin{quote}
\small
Fly the robot to viewpoint(s) from which the outcome---and the likely cause of any failure---is clearly visible, and CAPTURE images there with get\_observation(note="...").
[\ldots]
This code is NOT graded, and the environment is RESET to its initial state before the next attempt---nothing you do here persists. Because nothing persists, you are not limited to looking: when something you need to know is not visible from ANY viewpoint, you may physically ACT ON THE WORLD to reveal it, including changing whatever you believe is in the way. The reset undoes every such change, so the only thing that survives this step is what you LEARN. Do not use this step to perform the task; use it to find out what you cannot otherwise know.
\end{quote}

\paragraph{Evidence reporting and reuse.}
The prompt instructs the agent to select useful views and explicitly record its findings:
\begin{quote}
\small
A bare get\_observation() is not shown again. Give note= to the views the next step will actually need: the outcome of each part you judged not settled, and whatever explains a failure or changes the next program.
[\ldots]
Your program's console output is carried verbatim into the next code-generation step.
\end{quote}
Images selected through \texttt{get\_observation(note=...)} are supplied with their notes and recorded end-effector positions at capture. Calls without \texttt{note} can support perception within the diagnostic program, but their images are not forwarded; measurements derived from them can be conveyed through console output. The selected images, notes, positions, diagnostic reasoning, and console output are supplied alongside the original task-execution feedback. In a single response, the agent either declares \texttt{FINISH} or returns \texttt{REGENERATE} with a complete revised task program.

\subsection{Feedback conditions and generator configurations}
\label{app:generators}

The primary comparison uses the same Claude Code--Opus~5 configuration across terminal feedback, execution traces, and AF. Terminal feedback contains the execution outputs and terminal observation; traces add records collected at primitive-call boundaries; AF adds observations and measurements acquired by a separately generated diagnostic program.

The cross-configuration study evaluates AF within each model--interface configuration before comparing configurations. Coding-agent and direct-API interfaces can differ in tool access, context management, and execution history, so cross-configuration differences are not interpreted as an isolated effect of the underlying language model.

\paragraph{Adaptation of CaP-X.}
Our implementation adapts CaP-X's robot APIs and fixed skill library to aerial manipulation, with the shared interfaces described in Appendix~A.4. We hold the model and agent interface fixed across the primary feedback comparison and do not use parallel ensembling. CaP-Agent0 uses a visual-differencing module to summarize changes between successive observations in text. In spatially extended, long-horizon aerial tasks, however, egocentric observations may cover different regions and omit outcomes needed to establish completion, limiting what such comparisons can reveal. We therefore leave visual interpretation to the coding agent by providing RGB observations directly. The reset protocol in Appendix~A.1 supports iterative refinement of a complete task program, with each revised program required to perform the entire task from the initial state.

\begin{figure}[t]
    \centering
    \includegraphics[width=0.95\linewidth]{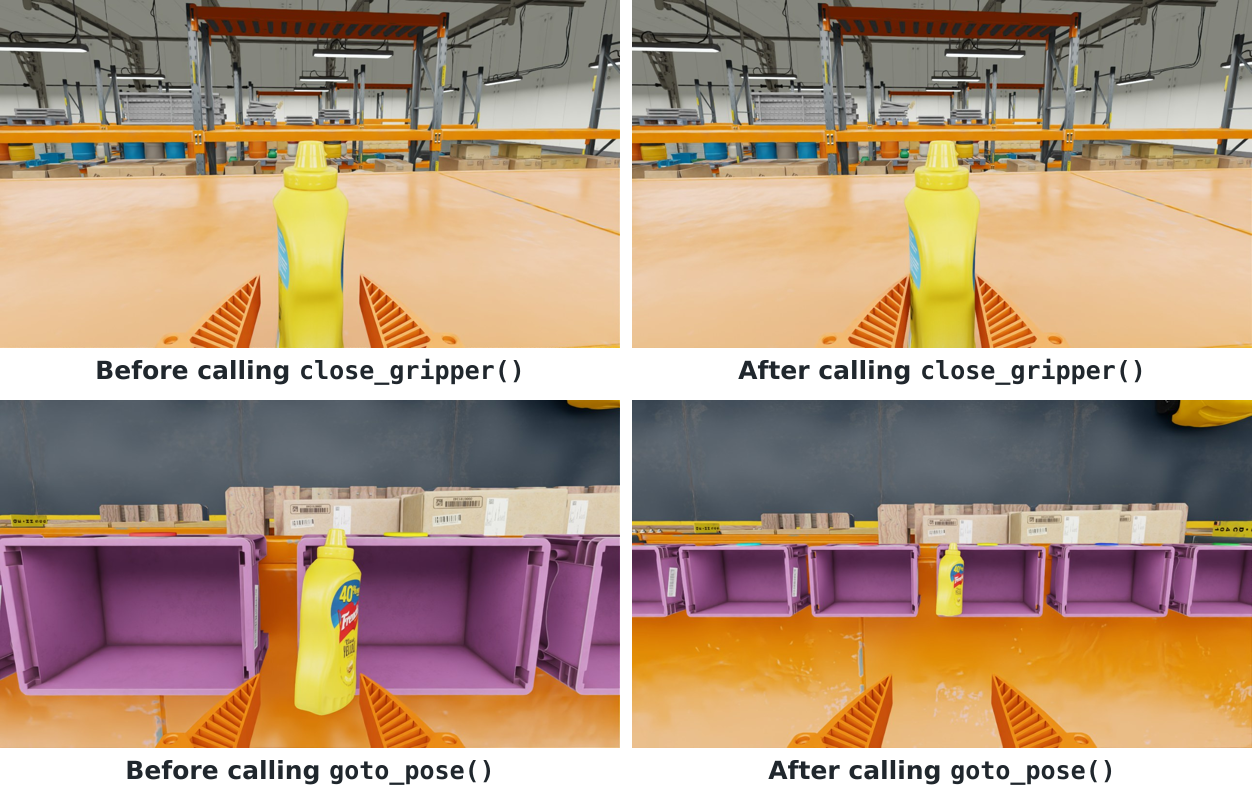}
    \caption{\textbf{Execution-trace feedback.} Images captured before and after a primitive call accompany its structured execution record. The observations follow the task execution rather than a dedicated verification trajectory.}
    \label{fig:trace}
\end{figure}

\paragraph{Execution traces.}
\label{app:trace}
Trace feedback pairs structured execution records with observations captured around primitive calls. The recorder follows ASPIRE's per-primitive tracing design~\citep{lu2026aspire}, adapted to our robot APIs and CaP-X-based loop. Before each post-execution \texttt{FINISH}/\texttt{REGENERATE} decision, the same coding agent is directed to the trace files and instructed to read the structured records and inspect relevant keyframes using file tools. Session instructions explain the record format and provide interpretation heuristics for execution signals. The recorded viewpoints follow the task program rather than a separately generated diagnostic objective. Figure~\ref{fig:trace} illustrates this interface.

\subsection{Simulation environments and robot configuration}
\label{app:system}

\begin{figure}[t]
    \centering
    \includegraphics[width=\linewidth]{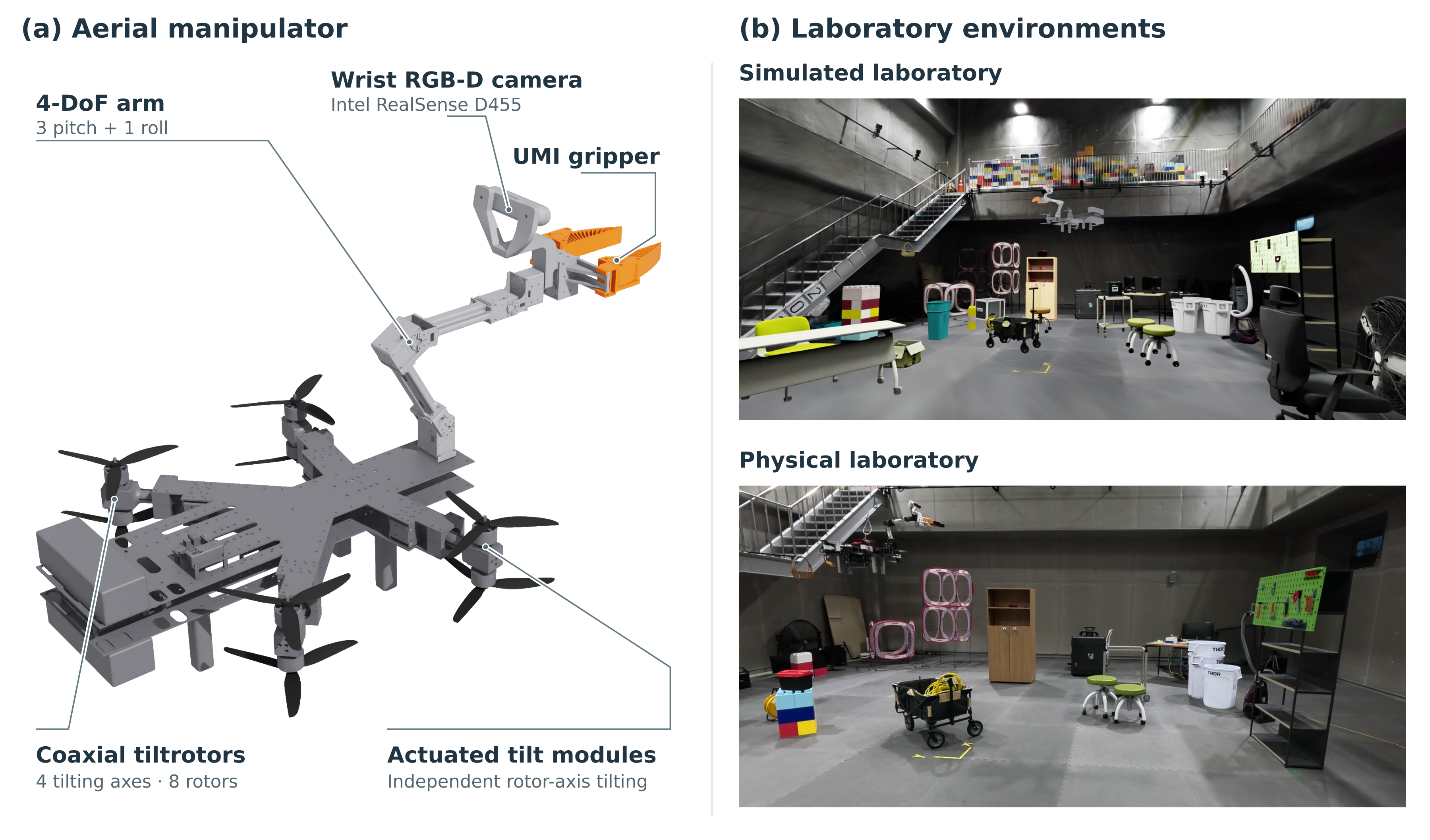}
    \caption{
    \textbf{Aerial manipulator and laboratory environment.}
    (a) Simulated model of our custom-designed platform, with its principal components annotated.
    (b) The laboratory digital twin and the physical laboratory, showing corresponding structures and task objects despite differences in their arrangement.
    }
    \label{fig:robot-laboratory-configuration}
\end{figure}

We build on AM-Bench~\citep{wang2026ambench}, a modular aerial-manipulation simulator that couples floating-base and manipulator dynamics with configurable low-level controllers.
We integrate two task environments and a model of our custom-designed omnidirectional aerial manipulator, using the same robot configuration and control architecture as our physical platform.
This setup allows synthesized programs to be evaluated under simulated flight dynamics and closed-loop tracking, rather than assuming perfect execution of commanded motions.
Figure~\ref{fig:robot-laboratory-configuration} presents the robot configuration and compares the simulated and physical laboratories.

\paragraph{Laboratory digital twin.}
We construct the laboratory digital twin from photographs and videos captured with an iPhone.
Individual object assets are generated using SAM 3D Objects~\citep{sam3dteam2025sam3d3dfyimages} and scaled to match directly measured dimensions of the corresponding physical objects.
For the laboratory background, we build on CoMe~\citep{radl2026come} and modify the 3D Gaussian rendering pipeline for integration with Isaac Sim.
During reconstruction, surface normals predicted by Metric3D v2~\citep{hu2024metric3dv2} and depth maps predicted by MVSAnywhere~\citep{izquierdo2025mvsanywhere} provide geometric supervision.
We assemble the reconstructed background and object assets into a task environment that preserves the principal laboratory structures and task objects, without requiring an exact match to the physical obstacle arrangement (Figure~\ref{fig:robot-laboratory-configuration}b).

\paragraph{Warehouse.}
We build the warehouse environment using NVIDIA Isaac Sim's \emph{Full Warehouse} asset~\citep{nvidia_isaac_sim_warehouse}.
Task objects are drawn from the assets provided with Isaac Sim, including a YCB mustard bottle, KLT totes, and cardboard boxes from the warehouse scene~\citep{nvidia_isaac_sim_assets}.
We configure their placement and task-specific visual markings to construct the evaluation scenarios.

\paragraph{Aerial manipulator.}
The simulated robot follows the configuration of our physical platform: a coaxial quad-tiltrotor with eight rotors, four actuated tilt joints, a four-degree-of-freedom arm, and a parallel-jaw UMI gripper (Figure~\ref{fig:robot-laboratory-configuration}a).
The arm comprises three pitch joints and one wrist-roll joint.
A wrist-mounted RGB-D camera provides egocentric observations whose viewpoint changes with flight and arm motion.
The robot operates under simulated flight dynamics through the shared execution backend described in the following subsection.

\subsection{Shared robot APIs and execution backend}
\label{app:apis}

Task programs and AF diagnostic programs use the same robot API surface.
Table~\ref{tab:robot-apis} summarizes the observation, perception, motion, and manipulation interfaces available in the final aerial-manipulation experiments.
Table~\ref{tab:geometry-utilities} lists the accompanying geometric and numerical utilities.
These tables describe the available functions; individual programs may use only a subset.

\begin{table}[p]
    \centering
    \small
    \setlength{\tabcolsep}{4pt}
    \renewcommand{\arraystretch}{1.08}

    \caption{\textbf{Robot APIs available to generated programs.}}
    \label{tab:robot-apis}
    \begin{tabularx}{\linewidth}{
        @{}
        >{\raggedright\arraybackslash}p{0.46\linewidth}
        >{\raggedright\arraybackslash}X
        @{}
    }
        \toprule
        API & Function \\
        \midrule
        \multicolumn{2}{@{}l}{\textbf{Observation and motion}} \\
        \texttt{get\_observation}
        & Retrieve wrist RGB-D images, camera parameters, and robot state. \\
        \texttt{get\_ee\_pose}
        & Read the end-effector pose and gripper opening. \\
        \texttt{goto\_pose}
        & Move the grasp center to a target position with a specified approach direction and roll. \\
        \texttt{open\_gripper}, \texttt{close\_gripper}
        & Open or close the gripper. \\
        \midrule
        \multicolumn{2}{@{}l}{\textbf{Image-based perception}} \\
        \texttt{point\_prompt\_molmo}
        & Locate a language-specified target in image coordinates. \\
        \texttt{segment\_sam3\_text\_prompt}
        & Segment targets specified by a text prompt. \\
        \texttt{segment\_sam3\_point\_prompt}
        & Segment a target specified by an image point. \\
        \midrule
        \multicolumn{2}{@{}l}{\textbf{Semantic-map queries}} \\
        \texttt{find\_objects}
        & Retrieve mapped objects matching a label. \\
        \texttt{nearest\_object}
        & Find the mapped object nearest a world-frame point. \\
        \texttt{object\_at\_point}
        & Query the mapped object occupying a world-frame point. \\
        \texttt{describe\_semantic\_map}
        & Summarize the currently mapped objects as text. \\
        \midrule
        \multicolumn{2}{@{}l}{\textbf{Grasp proposal and selection}} \\
        \texttt{plan\_grasp}
        & Propose scored grasps from depth, intrinsics, and segmentation. \\
        \texttt{plan\_grasp\_from\_point\_clouds}
        & Propose scored grasps from scene and target point clouds. \\
        \texttt{select\_horizontal\_grasp}
        & Select a candidate with a near-horizontal approach. \\
        \bottomrule
    \end{tabularx}

    \vspace{1em}

    \caption{
    \textbf{Geometric and numerical utilities.}
    }
    \label{tab:geometry-utilities}
    \begin{tabularx}{\linewidth}{
        @{}
        >{\raggedright\arraybackslash}p{0.46\linewidth}
        >{\raggedright\arraybackslash}X
        @{}
    }
        \toprule
        Utility & Function \\
        \midrule
        \multicolumn{2}{@{}l}{\textbf{Depth and point-cloud processing}} \\
        \texttt{depth\_to\_point\_cloud}
        & Deproject depth into camera-frame 3D points. \\
        \texttt{mask\_to\_world\_points}
        & Deproject masked depth into world-frame points. \\
        \texttt{pixel\_to\_world\_point}
        & Convert a pixel and depth into a world-frame point. \\
        \texttt{get\_oriented\_bounding\_box\_\allowbreak{}from\_3d\_points}
        & Estimate an oriented bounding box from 3D points. \\
        \texttt{filter\_noise}
        & Remove point-cloud outliers. \\
        \texttt{subsample\_point\_cloud}
        & Reduce the number of points in a point cloud. \\
        \midrule
        \multicolumn{2}{@{}l}{\textbf{Coordinate transformations and geometry}} \\
        \texttt{rotation\_matrix\_to\_quaternion}
        & Convert a rotation matrix into a quaternion. \\
        \texttt{decompose\_transform}
        & Extract position and quaternion from a transform. \\
        \texttt{transform\_points}
        & Apply a homogeneous transform to 3D points. \\
        \texttt{interpolate\_segment}
        & Generate points along a straight line segment. \\
        \texttt{normalize\_vector}
        & Normalize a nonzero vector to unit length. \\
        \bottomrule
    \end{tabularx}
\end{table}

\paragraph{Composing perception and action.}
The API exposes separate operations rather than a single object-to-action routine.
For example, a program can query the semantic map for candidate locations, move to an informative viewpoint, segment the target, recover its geometry from depth, and command an approach followed by gripper closure.
Grasp-proposal functions return candidates rather than executing grasps.

\paragraph{Semantic-map APIs.}
We maintain an object-level 3D semantic map to provide spatial references for objects outside the current camera view.
The perception pipeline uses OWLv2~\citep{minderer2024OWLv2} for open-vocabulary 2D detection and Boxer~\citep{detone2026boxer} for 3D box estimation and multi-view integration using RGB-D observations and camera poses.
The detection vocabulary is fixed by the task configuration in the final experiments.
Our mapping layer adds sensor-range weighting and distinguishes depth-supported estimates from RGB-only hypotheses, which are treated as directional cues.
The map exposes observation-derived labels and geometry estimates through the queries in Table~\ref{tab:robot-apis}; it contains no ground-truth object poses.
Because labels can be incorrect and locations can become erroneous, the agent is instructed to use map entries as navigation and reinspection cues and verify them through direct perception.
The map is cleared together with the environment before the next task-program execution.

\paragraph{Collision-aware motion planning.}
\texttt{goto\_pose} executes a motion only after a collision-free whole-body trajectory has been found and verified.
The workspace is represented by a voxel occupancy map converted to a signed distance field, and the robot by 30 collision spheres on the airframe, arm, and gripper.
Goal configurations for the airframe and the four arm joints are solved by nonlinear least-squares inverse kinematics in PyRoki~\citep{kim2025pyroki}, and a route between them is found by A* search~\citep{hart1968formal} over airframe position, heading, and arm posture.
The resulting path is smoothed with a JAX-based optimizer~\citep{jax2018github} and time-parameterized as a quintic B\'ezier trajectory under velocity and acceleration limits, then tracked by a reference governor that slows the reference when tracking error grows and aborts if it leaves the verified safety margin.

\paragraph{Shared interface, environment-specific execution.}
The same task-level function names and calling conventions are used in simulation and physical deployment.
Observation calls obtain rendered sensor data and simulated state in simulation, and measured sensor data and estimated state on hardware.
For \texttt{goto\_pose}, the planner and reference governor above are shared, while the control stack that tracks the reference is specific to each environment.
The call reports whether the target was reached, the motion was refused with a stated reason, or execution timed out.
Optional arguments specify a final approach offset, settling behavior, and intended contact.
This separation preserves the generated program's perception and action interface while allowing the underlying sensing, calibration, and control implementations to differ.

\paragraph{Use during active feedback.}
AF does not introduce separate \texttt{view} or \texttt{probe} primitives.
Instead, a diagnostic program composes the same observation and action APIs to acquire additional views or conduct physical tests in simulation.
Execution-trace conditions retain the same callable API surface and add recording around API calls.
Thus, feedback conditions differ in how evidence is acquired and exposed to the coding agent, rather than in the robot functions available to the generated program.

\section{Tasks and Evaluation Protocol}
\label{app:evaluation}
\subsection{Task instructions and setup}
\label{app:tasks}

The primary evaluation contains six aerial-manipulation tasks. 1-Object Pick-and-place, 2-Objects Pick-and-place, Cabinet, and Payload use the laboratory digital twin; Sort and Align use the warehouse simulation.
Each box presents a task-instruction excerpt, with omissions marked by [\ldots]. The accompanying setup descriptions provide experimental context and are not part of the quoted instructions.

\setcounter{instructiontask}{0}

\taskentry
{1-Object Pick-and-place}
{Pick the hammer off the pegboard and put it in the basket hanging on the stairs.}
{The hammer is mounted among other tools on a pegboard. The destination basket is across the hall, outside the source view.}

\taskentry
{2-Objects Pick-and-place}
{Put the hammer in the basket on the stairs, and put the wrench on the upper shelf in the open section of the cabinet across the hall.}
{The two tools share the pegboard source but have spatially separated destinations.}

\taskentry
{Cabinet}
{Take the water bottle from the cabinet across the hall and put it in the basket hanging on the stairs.}
{The bottle initially sits behind two closed, outward-opening cabinet doors, which must be opened sufficiently to allow access and extraction.}

\taskentry
{Payload}
{Eight water bottles stand on the long brown table; they differ only in colour. Three of them are empty; the other five hold water, and how much they hold may differ. Move the five bottles that hold water and both wicker baskets onto the long white table across the hall, and leave the empty bottles on the brown table. The mission is not only to get them there but to do it with the fewest crossings of the hall: a crossing is any flight between the two tables' sides of the hall, in either direction, and every crossing counts. [\ldots]}
{Bottle masses and the carrying limit are not disclosed in the initial instruction. The environment prevents simulated grasp slippage from causing transport failures within the payload limit. Motion commands are rejected when the carried load exceeds the payload limit. This setup more faithfully reflects the carrying behavior observed on our physical platform and is applied identically across all comparison conditions.}

\taskentry{Sort}
{Pick the mustard bottle off the rack shelf and put it in the matching tote. PURPLE KLT totes (plastic bins) stand in a row on a rack deck; each tote wears ONE round coloured sticker. The bottle MUST go into the tote whose sticker has the SAME colour as the object you pick up [\ldots].}
{The bottle starts on a high rack shelf, while nine purple totes stand on another rack. Their colored stickers distinguish the target tote from the distractors.}

\taskentry{Align}
{Some of the cartons in this bay carry a YELLOW STICKER, and any of them that sticks out past the shelf front into the aisle is in danger of falling. Push every such carton in so that its front face sits flush with, and parallel to, the front faces of the neighbouring cartons on the same shelf -- as flush and as parallel as you can make it. [\ldots]}
{Two target cartons protrude from the rack. Their centers of mass are offset from their geometric centers, a property neither visible from their appearance nor disclosed in the instruction. A push at the geometric center can therefore induce rotation.}

\subsection{Task completion and progress}
\label{app:scoredefs}

\paragraph{Final task completion.}
A trial is counted as successful only if the final task-program execution satisfies the task-specific completion criteria, evaluated before Active Feedback (AF), and the agent declares \texttt{FINISH}. A false \texttt{FINISH} remains a failure.

For feedback condition $a$, task $\tau$, and trial $i$, let $y_{a,\tau,i}$ denote final task success. We run ten trials for each of the six primary tasks under each condition and give each task equal weight:
\begin{equation}
    \bar{C}_{a}
    =
    \frac{1}{|\mathcal{T}|}
    \sum_{\tau\in\mathcal{T}}
    \frac{1}{n_{\tau}}
    \sum_{i=1}^{n_{\tau}} y_{a,\tau,i},
    \qquad n_{\tau}=10.
    \label{eq:completion}
\end{equation}
Because trial counts are equal across tasks, this task-macro-average equals the pooled success rate.

\paragraph{Progress scores.}
We normalize each task's progress score by its rubric maximum: 11 for Cabinet, 5 for 1-Object Pick-and-place, 9 for 2-Objects Pick-and-place, 8 for Payload, 6 for Sort, and 2 for Align.
Progress is computed from task-program execution records; physical progress obtained during AF is not credited.
A correct \texttt{FINISH} declaration earns the declaration-related point only when the task's completion conditions are satisfied.
An incorrect declaration earns no declaration-related point.
Trials that terminate early retain their final progress score through turn 11.

\paragraph{Cabinet (maximum: 11).}
Each stage contributes one point. Reaching a later stage credits the preceding stages.
The door-opening threshold is $55^\circ$, determined by testing cabinet access in $1^\circ$ increments; this was the minimum opening angle that allowed the drone's manipulator to enter the cabinet in our setup.

\begin{center}
\begin{tabularx}{\linewidth}{@{}cX@{}}
\toprule
Points & Criterion \\
\midrule
1 & A cabinet door handle is observed. \\
1 & A door handle is grasped. \\
1 & The grasped door is moved. \\
1 & One door is opened by at least $55^\circ$. \\
1 & The other door handle is grasped with one door open. \\
1 & The other door is moved with one door open. \\
1 & Both doors are opened by at least $55^\circ$. \\
1 & The bottle is grasped. \\
1 & The bottle is extracted from the cabinet. \\
1 & The bottle is placed inside the destination basket. \\
1 & The agent declares \texttt{FINISH} with the bottle successfully placed inside the basket. \\
\bottomrule
\end{tabularx}
\end{center}

\paragraph{1-Object Pick-and-place (maximum: 5).}
The following items each contribute one point.
Placement records successful settling at the destination (hanging basket), whereas retention checks whether the object remains there at the end of the attempt.

\begin{center}
\begin{tabularx}{\linewidth}{@{}clX@{}}
\toprule
Points & Item & Criterion \\
\midrule
1 & Grasp & The hammer is held and lifted at least 2\,cm from its mount. \\
1 & Transport & The held hammer is brought within 0.5\,m of the destination center. \\
1 & Placement & The hammer is placed at the destination and remains there for at least 0.5\,s. \\
1 & Retention & The hammer remains at the destination at the end of the attempt. \\
1 & \texttt{FINISH} & The agent declares completion while the hammer remains successfully placed. \\
\bottomrule
\end{tabularx}
\end{center}

\paragraph{2-Objects Pick-and-place (maximum: 9).}
The hammer and wrench are scored separately: the hammer's destination is the hanging basket, and the wrench's destination is the designated shelf.
The item scores are summed, with one additional point for a correct completion declaration.

\begin{center}
\begin{tabularx}{\linewidth}{@{}clX@{}}
\toprule
Points & Item & Criterion \\
\midrule
1 & Hammer grasp & The hammer is held and lifted at least 2\,cm from its mount. \\
1 & Hammer transport & The held hammer is brought within 0.5\,m of its destination center. \\
1 & Hammer placement & The hammer is placed at its destination and remains there for at least 0.5\,s. \\
1 & Hammer retention & The hammer remains at its destination at the end of the attempt. \\
1 & Wrench grasp & The wrench is held and lifted at least 2\,cm from its mount. \\
1 & Wrench transport & The held wrench is brought within 0.5\,m of its destination center. \\
1 & Wrench placement & The wrench is placed at its destination and remains there for at least 0.5\,s. \\
1 & Wrench retention & The wrench remains at its destination at the end of the attempt. \\
1 & \texttt{FINISH} & The agent declares completion while both objects remain successfully placed. \\
\bottomrule
\end{tabularx}
\end{center}

\paragraph{Payload (maximum: 8).}
The task contains five filled bottles and three empty bottles.
The rubric recognizes two transport pairs: red--black and yellow--blue.
The green bottle is transported separately.
Each item below contributes one point.

\begin{center}
\begin{tabularx}{\linewidth}{@{}clX@{}}
\toprule
Points & Item & Criterion \\
\midrule
1 & Loading & A filled bottle is loaded into a basket. \\
1 & Valid pair & A basket contains exactly one valid pair: red--black or yellow--blue. \\
1 & Both pairs & Both valid pairs are loaded into their respective baskets. \\
1 & First pair delivered & One valid pair is delivered in its basket to the destination table. \\
1 & Both pairs delivered & Both valid pairs are delivered in their baskets to the destination table. \\
1 & Green bottle delivered & The green bottle is carried separately to the destination table. \\
1 & Empty bottles preserved & All three empty bottles remain at the source table. \\
1 & \texttt{FINISH} & The agent declares completion with all five filled bottles and both baskets at the destination, all three empty bottles at the source, and no more than six crossings. \\
\bottomrule
\end{tabularx}
\end{center}

The empty-bottle criterion is already satisfied initially and is lost if the required source configuration is not preserved.

\paragraph{Sort (maximum: 6).}
The target tote is the one whose sticker matches the bottle's color.
Each item below contributes one point.

\begin{center}
\begin{tabularx}{\linewidth}{@{}clX@{}}
\toprule
Points & Item & Criterion \\
\midrule
1 & Grasp & The bottle is grasped. \\
1 & Extraction & The bottle is moved out of the source shelf. \\
1 & Transport & The held bottle is positioned above the target tote. \\
1 & Release & Above the target tote, the gripper opening reaches at least 75\,mm while the bottle is within 0.25\,m vertically of the tote rim. \\
1 & Retention & The task evaluator confirms successful placement in the target tote. \\
1 & \texttt{FINISH} & The agent declares completion while the bottle is successfully placed in the target tote. \\
\bottomrule
\end{tabularx}
\end{center}

\paragraph{Align (maximum: 2).}
The task requires aligning two protruding cartons marked with yellow stickers with the neighboring carton fronts.
Let $e_{\max}$ be the largest absolute perpendicular distance from any evaluated front corner of either target carton to the reference plane.
The precision component is
\begin{equation}
p =
\operatorname{clip}\left(
1-\frac{e_{\max}}{15\,\mathrm{cm}},\,0,\,1
\right).
\end{equation}
The 15\,cm normalization range follows the supplied scoring protocol and was selected using the CaP-X final-error median of approximately 15.1\,cm.
The two components below are added, and their sum is divided by 2 for the progress curves.

\begin{center}
\begin{tabularx}{\linewidth}{@{}clX@{}}
\toprule
Points & Item & Criterion \\
\midrule
$0$--$1$ & Precision & The continuous precision score $p$ defined above. \\
1 & \texttt{FINISH} & The agent declares completion with $e_{\max}\leq 5\,\mathrm{cm}$. \\
\bottomrule
\end{tabularx}
\end{center}

For the reported Align results, a true positive requires a \texttt{FINISH} declaration with $e_{\max}\leq 5\,\mathrm{cm}$.

For normalized score $r_{a,\tau,i,t}$ and final execution index $T_{a,\tau,i}$, the standing-score curve is
\begin{equation}
    S_{a,\tau}(t)
    =
    \frac{1}{n_{\tau}}
    \sum_{i=1}^{n_{\tau}}
    r_{a,\tau,i,\min(t,T_{a,\tau,i})}.
    \label{eq:standing}
\end{equation}
Stopped trials retain their final score. The standing score can decrease after a revision, whereas a running maximum retains the best score reached by any preceding execution, regardless of whether the agent kept that program. These quantities answer different questions and are not used interchangeably.

\subsection{Completion judgments and statistical analysis}

Completion-judgment precision is the number of successful terminal \texttt{FINISH} declarations divided by the total number of terminal \texttt{FINISH} declarations. Trials ending in true \texttt{FINISH}, false \texttt{FINISH}, and exhaustion of the execution limit without \texttt{FINISH} are reported separately. These trial-level outcomes are distinct from error rates over all intermediate completion decisions.

Progress curves use ten trials per task and condition. Shaded bands are pointwise 95\% percentile-bootstrap intervals obtained from 1,000 trial-level resamples: within each task and
condition for task-specific curves, and over the pooled 60 trials within each condition for ALL TASKS. Repeated task-program executions within a trial are not treated as independent samples, and matching trial indices across conditions are not assumed to define paired observations. Reported comparisons characterize the evaluated task suite.

\subsection{Information access}
\label{app:integrity}

The agent receives observations and execution outputs through the robot APIs and the feedback interface specified for each condition.
Evaluator-only ground-truth states, task scores, and success labels are excluded from this feedback; grader logs are recorded separately from agent-visible execution output.
The agent therefore makes completion decisions from the available execution evidence rather than receiving the evaluator's verdict.

\section{Additional Experimental Results}
\label{app:results}
\subsection{Task progress in the main evaluation}
\label{app:task-progress}

\begin{figure}[t]
  \centering
  \includegraphics[width=\linewidth]{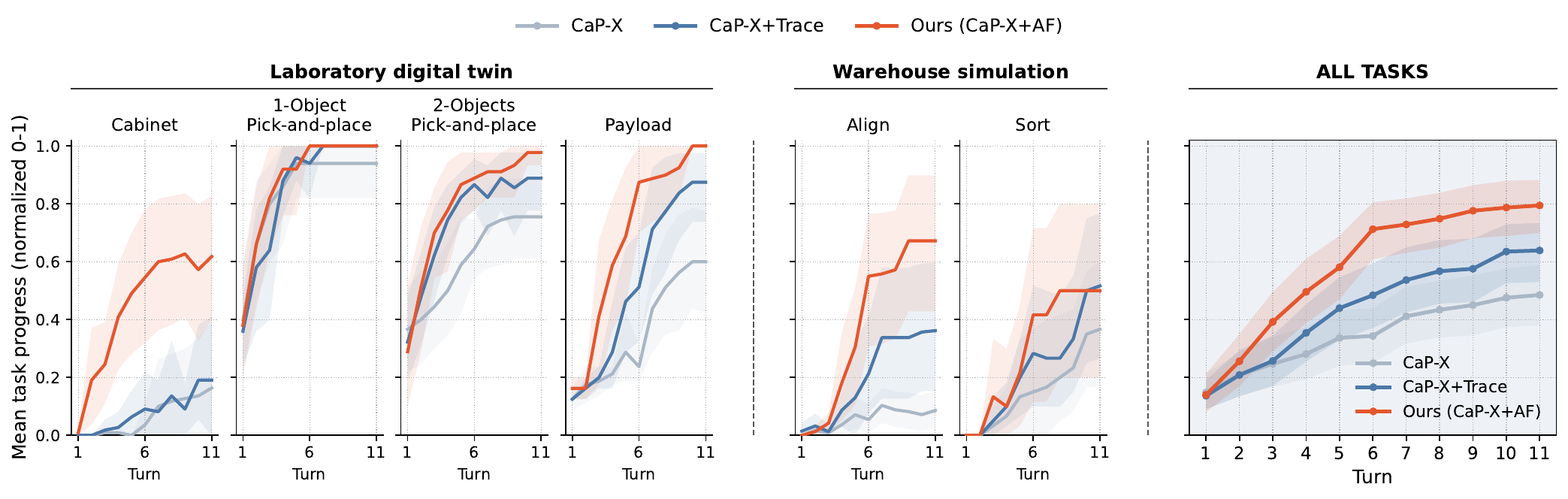}
  \caption{Task progress across task-program executions. AF (Ours) achieves generally comparable or higher progress across individual tasks and shows a clear advantage in average progress across all six tasks. Turn 1 denotes the initial program execution, and trials that terminate early retain their final score through turn 11. Scores are normalized by the maximum score of each task-specific rubric (Appendix~\ref{app:scoredefs}).}
  \label{fig:task-progress}
\end{figure}

Figure~\ref{fig:task-progress} shows task progress across successive task-program executions for the three feedback methods on all six tasks.

\subsection{Task progress across models and interfaces}
\label{app:model-interface-results}

\begin{figure}[htbp]
  \centering
  \includegraphics[width=\linewidth]{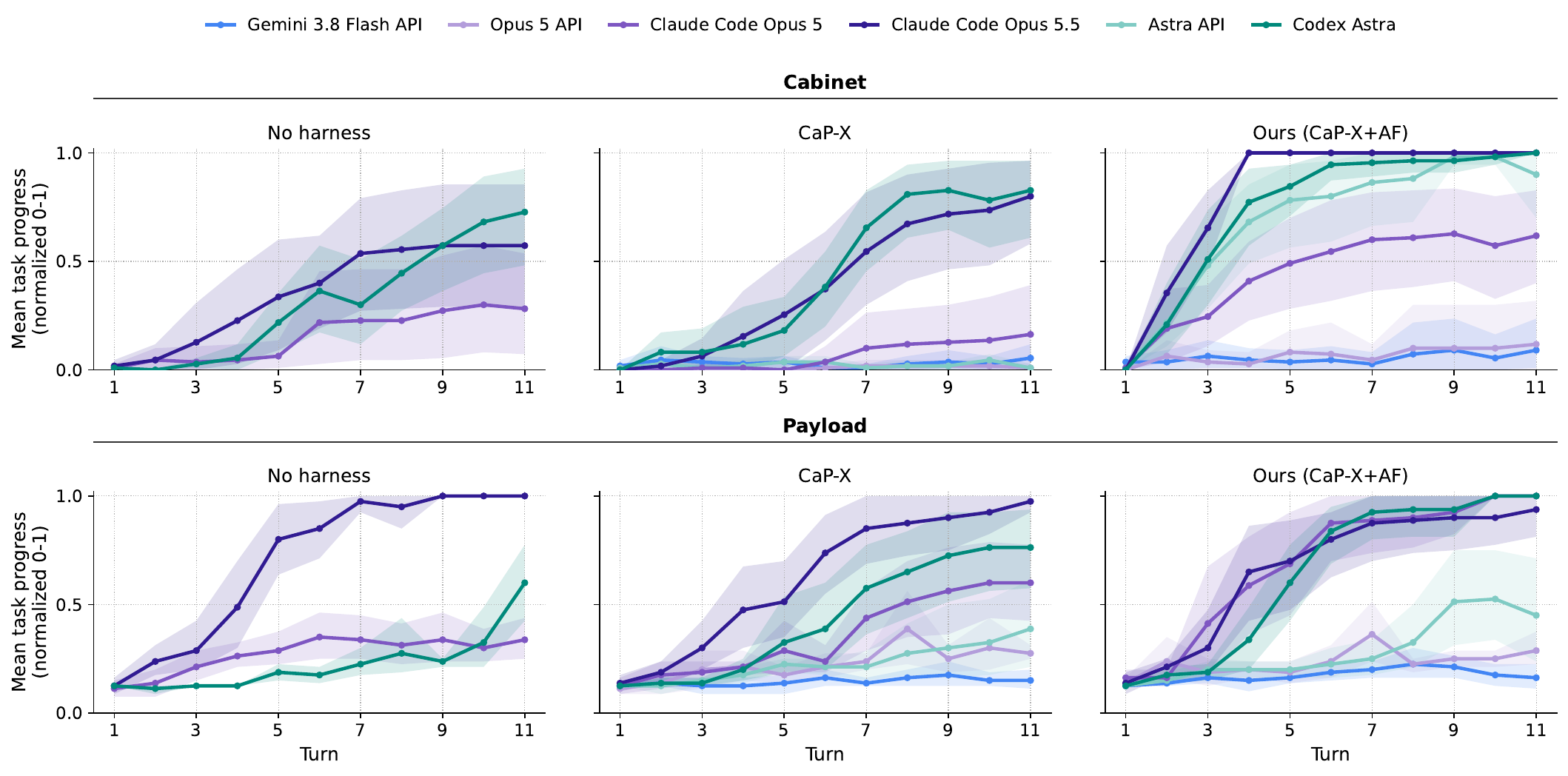}
  \caption{Task progress across API and coding-agent configurations. Each panel compares configurations under the same feedback method. Astra API achieves higher final average progress with AF than with CaP-X on both tasks, with a larger improvement on Cabinet. Normalization and early-termination handling follow Figure~\ref{fig:task-progress}.}
  \label{fig:api-ablation-progress}
\end{figure}

Figure~\ref{fig:api-ablation-progress} shows task progress across models and agent interfaces, including Gemini 3.8 Flash API~\citep{google2026gemini38flash}, with rows corresponding to tasks and columns to feedback methods.

\subsection{Semantic map ablation}
\label{app:map-ablation}
\begin{table}[htbp]
    \centering
    \caption{
        Semantic map ablations within AF, with ten trials per task
        and condition. Success counts successful trials; FP counts
        false \texttt{FINISH} declarations. Cost is the mean estimated
        model cost per trial in USD.
    }
    \label{tab:map-ablation}
    \begin{minipage}{0.65\linewidth}
        \centering
        \input{tables/map_ablation.tex}
    \end{minipage}
\end{table}
\paragraph{Role of semantic map support.}
We ablate semantic map support within Active Feedback (AF) on \textit{2-Objects Pick-and-place} and \textit{Sort}, which involve spatially separated task areas. We remove semantic map APIs, the semantic map summary included in the AF prompt, and semantic map-use instructions, while retaining the remaining perception and control APIs and the AF diagnostic stage. Task success drops from 9/10 to 4/10 on 2-Objects Pick-and-place and from 5/10 to 2/10 on Sort, while false \texttt{FINISH} declarations increase from 1 to 5 and from 0 to 3, respectively (Table~\ref{tab:map-ablation}).

Execution records suggest that the semantic map supports spatial search during task execution. Aerial mobility enables observations from diverse viewpoints, but selecting informative viewpoints remains a separate decision. Program-invoked perception tools process selected views and queries, whereas the background mapper accumulates detections during flight, providing cues about objects outside the current view. On the first execution of 2-Objects Pick-and-place, both conditions reached a hammer grasp in 5/10 trials. Without the map, however, distant basket localization became less accurate and mean SAM3 calls per task-program execution increased from 14 to 54, suggesting greater effort to reconstruct spatial information.

During AF, accumulated detections can also identify candidates for further inspection. All three false \texttt{FINISH} declarations in Sort without map support followed a final AF step that skipped observation and interpreted the target objects as absent. In one trial with map support, repeated SAM3 queries failed to detect the targets, but the map supplied target-tote candidates and the agent continued to \texttt{REGENERATE}. This suggests that map information can encourage continued search when individual perception queries provide no detections. Mean estimated model cost per trial increased from \$10.2 to \$15.3 on 2-Objects Pick-and-place and from \$29.3 to \$37.6 on Sort without map support. These costs reflect both execution count and termination decisions and should therefore be interpreted alongside success and false \texttt{FINISH} counts.

The map provides inspection targets rather than definitive task completion evidence. The agent is warned about inaccurate labels and locations and instructed to verify them through close-range observations. Across 11 analyzed executions involving wrench loss, the map never directly reported the fallen wrench's new location; in the eight executions where the loss was recognized, AF images and close-range perception results provided the evidence. Incorrect labels and stale locations sometimes confused the assessment. Together, these observations illustrate complementary roles: the map guides what to recheck and where, while AF checks the current state. Because map support is removed throughout the loop, this ablation does not isolate its contributions to task execution and diagnosis.

\subsection{Resource use}
\label{app:resources}

Table~\ref{tab:resource-usage} reports resource usage for the main evaluation (ten trials per task and method).
Model usage includes initial generation, subsequent decision calls, and AF diagnostic calls where applicable.
Per-success values equal total resource usage across all trials divided by the number of successes.

\begin{table}[htbp]
  \centering
  \caption{Resource usage in the main evaluation.
  Turns denote task-program executions; output tokens include thinking tokens
  and are reported in thousands (k).
  USD denotes estimated token-based LLM cost, not actual subscription charges.
  Per-success values include resources spent on failed trials;
  \mbox{--} denotes no observed success.
  Bold indicates the highest success count and lowest resource per
  success within each task block.}
  \label{tab:resource-usage}
  \setlength{\tabcolsep}{3.5pt}
  \resizebox{\linewidth}{!}{\input{tables/resource_usage}}
\end{table}

\clearpage
\section{Diagnostic Programs and Case Studies}
\label{app:cases}
\label{app:failure-audit}

The case studies examine the evidence available after task execution
and how it informs completion assessment and program revision.
They illustrate observation gaps in execution feedback and how AF
acquires additional evidence to investigate unresolved outcomes.

\subsection{Evidence for completion assessment}

We present two separate trials from object pick-and-place tasks.
The first illustrates a false completion declaration under CaP-X+Trace,
where the inspected keyframes did not reveal whether the object
had reached its destination.
The second shows how Active Feedback (AF) acquired new viewpoints to identify
unsuccessful placements despite the task program's success reports.

\input{sections/case_study/k1_trace_case}
\input{sections/case_study/k2_case}

\subsection{Evidence for program revision}
\label{app:align_case}

We present two cases in which AF-generated physical experiments
inform program revision. In Align, the agent tests and compares
push actions to revise contact selection. In Payload, it tests
a proposed basket load to revise the transport plan.

\input{sections/case_study/align_probe_case}
\input{sections/case_study/payload_case}

\section{Physical System and Deployment}
\label{app:hardware}

\subsection{Platform and execution system}
\label{app:hardware_platform}

\paragraph{Aerial platform.}
Our custom-designed omnidirectional aerial manipulator comprises a coaxial quad-tiltrotor with eight rotors, four actuated tilt joints, a four-degree-of-freedom arm, and a parallel-jaw UMI gripper.
The arm has three pitch joints and one wrist-roll joint.
A wrist-mounted RGB-D camera provides the egocentric visual observations used by the task program.
External cameras record the experiments for visualization and outcome assessment; their images are not supplied to the program.
Figure~\ref{fig:robot-laboratory-configuration} illustrates the robot configuration, and Table~\ref{tab:physical-platform-specifications} summarizes the physical system.

\begin{table}[t]
    \centering
    \small
    \setlength{\tabcolsep}{4pt}
    \renewcommand{\arraystretch}{1.12}
    \caption{\textbf{Physical platform specifications.}}
    \label{tab:physical-platform-specifications}
    \begin{tabularx}{\linewidth}{
        @{}
        >{\raggedright\arraybackslash}p{0.30\linewidth}
        >{\raggedright\arraybackslash}X
        @{}
    }
        \toprule
        Item & Specification \\
        \midrule
        \multicolumn{2}{@{}l}{\textbf{Airframe}} \\
        Configuration
        & Omnidirectional coaxial quad-tiltrotor: four tilting arms, each carrying a coaxial pair of rotors (eight rotors, four tilt joints). \\
        Dimensions
        & Airframe $0.62 \times 0.69$\,m, height $0.20$\,m including rotors. \\
        Mass
        & $8.30$\,kg in flight configuration (battery, onboard computer, camera, and gripper included). \\
        Motors
        & $8\times$ T-Motor VELOX V2812, 1155\,KV (6S). \\
        Propellers
        & $8\times$ T-Motor C7.5$\times$4.6 three-blade cinematic propellers (7.5\,in). \\
        Electronic speed controllers
        & $2\times$ HobbyWing XRotor FPV G2 4-in-1 ESC (65\,A). \\
        Battery
        & Two 6S1P semi-solid-state lithium-polymer packs (HYBRID PT-B18000-SS-6S), $22.2$\,V, $18{,}000$\,mAh. \\
        Tilt actuators
        & $4\times$ ROBOTIS DYNAMIXEL XM430-W210-T (one per arm). \\
        \midrule
        \multicolumn{2}{@{}l}{\textbf{Manipulator and gripper}} \\
        Arm
        & Four degrees of freedom (three pitch joints and one wrist-roll joint), actuated by $2\times$ DYNAMIXEL XM540-W270-R and $2\times$ DYNAMIXEL XM430-W210-R. \\
        Gripper
        & Parallel-jaw UMI gripper driven by one DYNAMIXEL XM430-W210-R. Symmetric left and right finger links are coupled through a mimic joint so that both fingers move evenly; maximum opening $80$\,mm. \\
        \midrule
        \multicolumn{2}{@{}l}{\textbf{Sensing and state estimation}} \\
        Wrist camera
        & Intel RealSense D455 RGB-D camera (active stereo depth, global-shutter RGB), operated at $1280\times720$ for both color and depth at $30$\,Hz with depth aligned to color. \\
        Onboard IMU
        & Holybro Pixhawk 6C flight controller with its built-in IMUs. \\
        External localization
        & OptiTrack motion capture (Motive) providing rigid-body pose at $100$\,Hz. \\
        State estimation
        & Extended Kalman filter fusing the onboard IMU with the external pose measurements. \\
        \midrule
        \multicolumn{2}{@{}l}{\textbf{Computation and control}} \\
        Onboard computer
        & ASUS NUC 15 Pro with an Intel Core~7 processor and $32$\,GB memory. \\
        Flight controller
        & Geometric nonlinear PID controller on SE(3)~\citep{goodarzi2013geometric}, running position and attitude control in a single $333$\,Hz ($3$\,ms) loop. \\
        Off-board perception
        & Workstation with an NVIDIA RTX A5000 GPU serving the perception and semantic-mapping models (Appendix~\ref{app:apis}). \\
        Communication
        & ROS~2 Humble (DDS) over Wi-Fi between the onboard computer, the ground laptop, and the perception workstation.\\
        \bottomrule
    \end{tabularx}
\end{table}

\subsection{Physical experiments and outcome assessment}
\label{app:hardware_trials}

\paragraph{Starting poses and scene differences.}
The robot's position and heading at task-program start, after takeoff, vary across trials.
Table~\ref{tab:hardware-start-poses} shows these starting poses for every evaluated trial.
The physical scene contains the corresponding task objects but does not exactly reproduce their simulated placement or the surrounding obstacle arrangement.
Table~\ref{tab:sim-real-object-positions} compares task-object positions in a common laboratory coordinate frame.
The digital twin and the motion-capture system share this frame, including its origin.

\begin{table}[t]
    \centering
    \small
    \setlength{\tabcolsep}{4pt}
    \renewcommand{\arraystretch}{1.12}
    \caption{\textbf{Scene layout in the laboratory digital twin and the physical laboratory.} Coordinates are in the shared world frame (meters). Simulation values are ground-truth poses from the scene description. Physical positions are motion-capture measurements at the closest corresponding point on each object, so the comparison is approximate rather than exact.}
    \label{tab:sim-real-object-positions}
    \begin{tabularx}{\linewidth}{
        @{}
        >{\raggedright\arraybackslash}X
        >{\raggedright\arraybackslash}p{0.30\linewidth}
        >{\raggedright\arraybackslash}p{0.30\linewidth}
        @{}
    }
        \toprule
        Item & Simulation $(x, y, z)$ & Physical $(x, y, z)$ \\
        \midrule
        \multicolumn{3}{@{}l}{\textbf{Target objects}} \\
        Hammer & $(0.78, -3.00, 1.68)$ & $(0.84, -2.91, 1.41)$ \\
        Wrench & $(0.96, -3.01, 1.41)$ & $(1.02, -2.91, 1.38)$ \\
        Water bottle & $(3.79, 0.98, 0.85)$ & $(3.29, 0.40, 0.83)$ \\
        \midrule
        \multicolumn{3}{@{}l}{\textbf{Task fixtures}} \\
        Pegboard front face ($y$) & $-3.05$ & $-2.95$ \\
        Hanging basket (center) & $(2.57, 4.01, 1.86)$ & $(2.22, 3.37, 2.01)$ \\
        Cabinet (center) & $(3.91, 0.97, 0.86)$ & $(3.45, 0.40, 0.91)$ \\
        Upper shelf top ($z$) & $1.42$ & $1.40$ \\
        Door knob (left) & $(3.68, 1.07, 0.98)$ & $(3.22, 0.49, 1.04)$ \\
        Door knob (right) & $(3.68, 0.87, 0.98)$ & $(3.22, 0.31, 1.03)$ \\
        \bottomrule
    \end{tabularx}
\end{table}

\begin{table}[t]
    \centering
    \small
    \setlength{\tabcolsep}{6pt}
    \renewcommand{\arraystretch}{1.12}
    \caption{\textbf{Starting poses of the hardware trials.} Airframe position and heading at task-program start after takeoff, in the shared world frame.}
    \label{tab:hardware-start-poses}
    \begin{tabular}{@{}llcc@{}}
        \toprule
        Task & Trial & Position $(x, y, z)$ [m] & Heading [$^\circ$] \\
        \midrule
        2-Objects Pick-and-place & 1 & $(-0.66, 0.00, 1.86)$ & $+1.8$ \\
        & 2 & $(-0.90, -0.01, 1.86)$ & $-0.3$ \\
        & 3 & $(-0.97, -0.04, 1.86)$ & $-0.1$ \\
        & 4 & $(-0.87, -0.04, 1.86)$ & $-0.6$ \\
        & 5 & $(-0.71, 0.33, 1.86)$ & $+41.3$ \\
        \midrule
        Cabinet & 1 & $(-1.19, -0.39, 1.86)$ & $+20.4$ \\
        & 2 & $(-0.91, -0.26, 1.86)$ & $+20.2$ \\
        & 3 & $(-1.26, -0.24, 1.86)$ & $+22.0$ \\
        & 4 & $(-1.56, 0.08, 1.86)$ & $+90.7$ \\
        & 5 & $(-1.08, -0.06, 1.84)$ & $+34.6$ \\
        \bottomrule
    \end{tabular}
\end{table}

\begin{figure}[t]
    \centering
    \includegraphics[width=\linewidth]{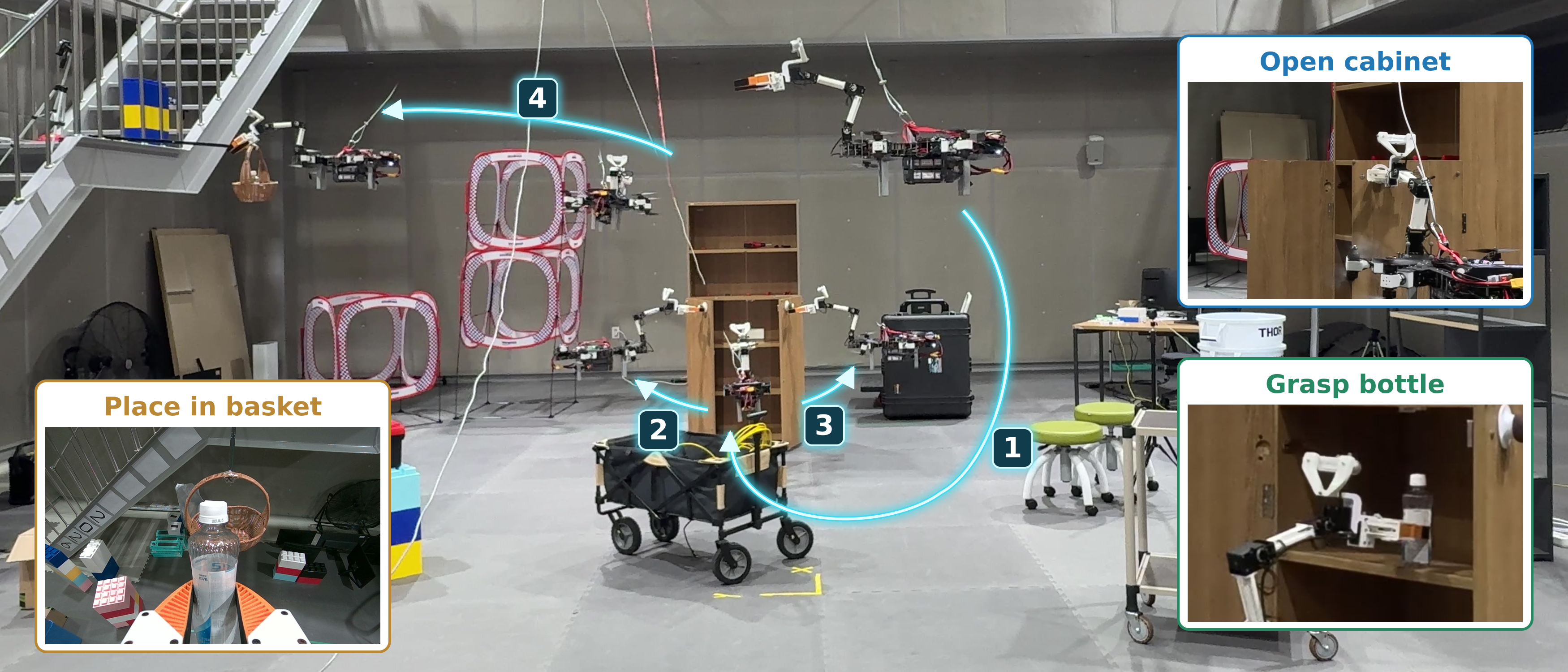}
    \caption{\textbf{Real-world Cabinet execution.} External views of one successful trial: opening the left door, opening the right door, grasping the bottle, and placing it in the hanging basket.}
    \label{fig:cabinet_success_stages}
\end{figure}

\begin{figure}[t]
    \centering
    \includegraphics[width=\linewidth]{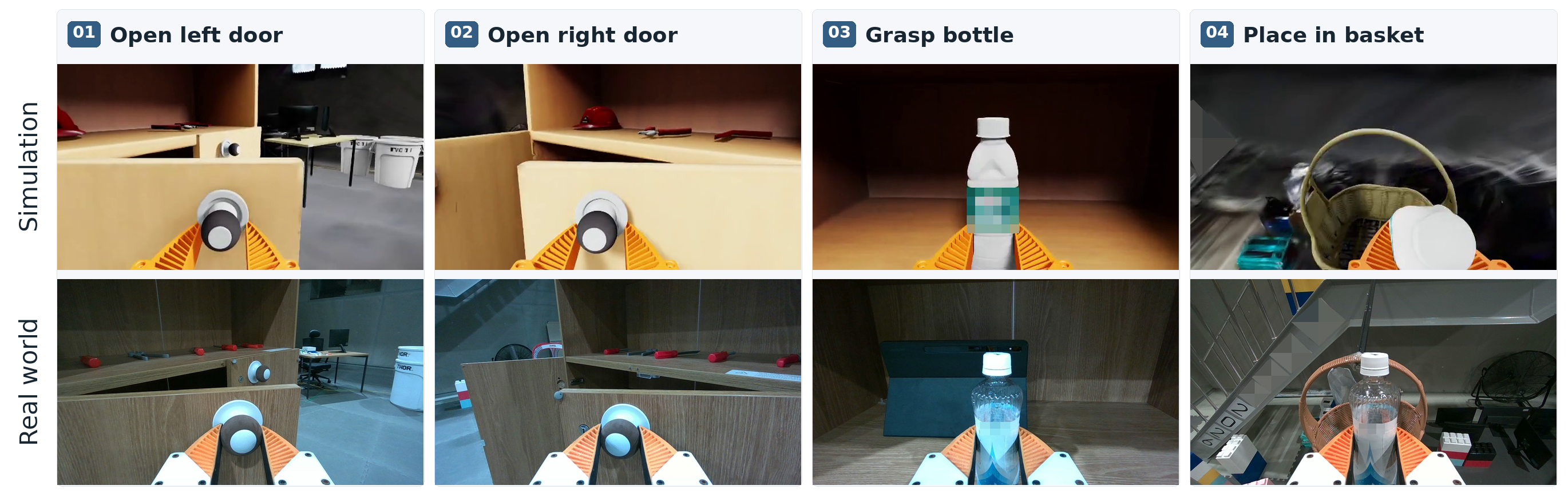}
    \caption{\textbf{Cabinet in simulation and the real world.} Wrist-camera views show opening the left door, opening the right door, grasping the bottle, and approaching the basket before release. The top and bottom rows show simulation and physical execution, respectively.}
    \label{fig:ee_sim_real_cabinet}
\end{figure}

\paragraph{Successful real-world Cabinet execution.}
Figure~\ref{fig:cabinet_success_stages} shows a successful real-world execution of the Cabinet task, complementing the 2-Objects Pick-and-place execution in Figure~\ref{k2_real_world_thumbnail}.
Figure~\ref{fig:ee_sim_real_cabinet} compares the same task in simulation and on the physical robot from the wrist camera.

\paragraph{Trial-specific outcomes.}
Table~\ref{tab:hardware-trial-details} lists the stage outcomes of every trial together with a short account of what happened.

\begin{table}[t]
    \centering
    \small
    \setlength{\tabcolsep}{3pt}
    \renewcommand{\arraystretch}{1.15}
    \caption{\textbf{Trial-level records of the real-world trials.} \checkmark: completed, $\times$: failed.}
    \label{tab:hardware-trial-details}
    \begin{tabularx}{\linewidth}{@{}c cccc >{\raggedright\arraybackslash}X@{}}
        \toprule
        Trial & S1 & S2 & S3 & S4 & What happened \\
        \midrule
        \multicolumn{6}{@{}l}{\textbf{2-Objects Pick-and-place} (S1 grasp hammer, S2 place in basket, S3 grasp wrench, S4 place on upper shelf)} \\
        1 & \checkmark & \checkmark & \checkmark & \checkmark &   \hfill --\hfill\null \\
        2 & \checkmark & \checkmark & \checkmark & \checkmark &  \hfill --\hfill\null \\
        3 & \checkmark & $\times$ & \checkmark & \checkmark & A perception error in the detected basket position shifted the release point, and the hammer struck the basket rim on release and fell outside. \\
        4 & \checkmark & \checkmark & \checkmark & \checkmark &  \hfill --\hfill\null \\
        5 & \checkmark & \checkmark & \checkmark & \checkmark &  \hfill --\hfill\null \\
        \midrule
        \multicolumn{6}{@{}l}{\textbf{Cabinet} (S1 open left door, S2 open right door, S3 grasp bottle, S4 place in basket)} \\
        1 & \checkmark & \checkmark & \checkmark & \checkmark &  \hfill --\hfill\null \\
        2 & \checkmark & \checkmark & \checkmark & \checkmark &  \hfill --\hfill\null \\
        3 & \checkmark & $\times$ & $\times$ & $\times$ & After releasing the left knob, the gripper became stuck on the handle and the robot could not move to the next waypoint; all subsequent motion calls were refused, and the trial failed. \\
        4 & \checkmark & \checkmark & \checkmark & $\times$ & A control error caused the gripper to grasp the lower part of the bottle, so the bottle occluded the wrist camera. The basket could therefore not be re-measured at close range, and the program failed to place the bottle inside it. \\
        5 & \checkmark & \checkmark & \checkmark & \checkmark &  \hfill --\hfill\null \\
        \bottomrule
    \end{tabularx}
\end{table}

%% file: tables/map_ablation.tex
\begingroup
\footnotesize
\setlength{\tabcolsep}{2pt}
\renewcommand{\arraystretch}{1.12}

\begin{tabularx}{\linewidth}{@{}Xccc@{}}
\toprule
AF variant & Success & FP & Cost (\$) \\
\midrule
\multicolumn{4}{@{}l}{\textit{2-Objects Pick-and-place}} \\
Full AF       & 9/10 & 1 & 10.2 \\
w/o semantic map       & 4/10 & 5 & 15.3 \\
\midrule
\multicolumn{4}{@{}l}{\textit{Sort}} \\
Full AF       & 5/10 & 0 & 29.3 \\
w/o semantic map       & 2/10 & 3 & 37.6 \\
\bottomrule
\end{tabularx}
\endgroup

%% file: tables/resource_usage.tex
\begin{tabular}{@{}llrrrrrrr@{}}
\toprule
& & & \multicolumn{3}{c}{Per trial} & \multicolumn{3}{c}{Per success} \\
\cmidrule(lr){4-6}\cmidrule(l){7-9}
Task & Method & Success & Turns & Output (k) & USD & Turns & Output (k) & USD \\
\midrule
All tasks & CaP-X & 19/60 & 7.2 & 142 & 9.6 & 22.7 & 447 & 30.4 \\
 & CaP-X+Trace & 29/60 & 7.0 & 129 & 9.4 & 14.6 & 267 & \textbf{19.5} \\
 & AF (Ours) & \textbf{45/60} & 6.0 & 155 & 17.1 & \textbf{8.1} & \textbf{206} & 22.8 \\
\midrule
Cabinet & CaP-X & 1/10 & 10.9 & 156 & 10.8 & 109.0 & 1562 & 107.7 \\
 & CaP-X+Trace & 0/10 & 10.9 & 174 & 13.7 & -- & -- & -- \\
 & AF (Ours) & \textbf{4/10} & 8.5 & 189 & 24.6 & \textbf{21.2} & \textbf{473} & \textbf{61.5} \\
\midrule
1-Object Pick-and-place & CaP-X & 9/10 & 2.4 & 27 & 1.8 & \textbf{2.7} & \textbf{30} & \textbf{2.0} \\
 & CaP-X+Trace & \textbf{10/10} & 3.6 & 53 & 3.4 & 3.6 & 53 & 3.4 \\
 & AF (Ours) & \textbf{10/10} & 2.9 & 68 & 5.4 & 2.9 & 68 & 5.4 \\
\midrule
2-Objects Pick-and-place & CaP-X & 4/10 & 5.5 & 104 & 6.1 & 13.8 & 259 & 15.4 \\
 & CaP-X+Trace & 6/10 & 4.8 & 105 & 7.1 & 8.0 & 175 & 11.8 \\
 & AF (Ours) & \textbf{9/10} & 4.7 & 111 & 10.2 & \textbf{5.2} & \textbf{123} & \textbf{11.3} \\
\midrule
Payload & CaP-X & 3/10 & 8.5 & 250 & 20.2 & 28.3 & 834 & 67.4 \\
 & CaP-X+Trace & 6/10 & 7.2 & 112 & 10.7 & 12.0 & 186 & 17.9 \\
 & AF (Ours) & \textbf{10/10} & 5.5 & 136 & 16.5 & \textbf{5.5} & \textbf{136} & \textbf{16.5} \\
\midrule
Align & CaP-X & 0/10 & 5.6 & 121 & 6.4 & -- & -- & -- \\
 & CaP-X+Trace & 3/10 & 6.6 & 160 & 8.8 & 22.0 & 534 & 29.4 \\
 & AF (Ours) & \textbf{7/10} & 6.3 & 203 & 16.7 & \textbf{9.0} & \textbf{290} & \textbf{23.8} \\
\midrule
Sort & CaP-X & 2/10 & 10.3 & 191 & 12.5 & 51.5 & 957 & 62.3 \\
 & CaP-X+Trace & 4/10 & 9.1 & 171 & 12.8 & 22.8 & \textbf{427} & \textbf{31.9} \\
 & AF (Ours) & \textbf{5/10} & 8.4 & 221 & 29.3 & \textbf{16.8} & 442 & 58.6 \\
\bottomrule
\end{tabular}

%% file: sections/case_study/k1_trace_case.tex
\paragraph{Missing outcome observations in execution traces.}
\label{app:k1-trace-evidence}

In a CaP-X+Trace trial, the task program attempted to place a hammer in a basket by moving to an estimated position above it, opening the gripper, and performing a visual check after moving upward. The following excerpts show the recorded execution signals and the observations available for completion assessment.

\begin{caseevidence}{1. Task-program output: approach and release without placement confirmation}
\begin{lstlisting}[style=casecode]
over basket: True
released; gripper: 1.0
[...]
hammer visible after release: None
\end{lstlisting}
\end{caseevidence}

The program set \texttt{over basket} to \texttt{True} when a motion to an estimated release position succeeded; this flag did not verify placement. After opening the gripper and moving upward, the program obtained no hammer detection. Neither the open gripper nor the absence of a detection established that the hammer had landed inside the basket.
The agent declared \texttt{FINISH}, although the evaluator recorded that the hammer-placement requirement was unmet.

The agent inspected release-related trace keyframes, including the two shown below, before declaring \texttt{FINISH}. These views showed the open gripper and only part of the basket rim at the lower image boundary, leaving placement inside the basket unverified. This case illustrates an observation gap in execution-trace feedback: recording approach and release does not necessarily provide evidence of the resulting placement.

\begin{caseevidence}{2. Inspected trace keyframes: placement remains unverified}
\noindent
\begin{minipage}[t]{0.485\linewidth}
\vspace{0pt}
\centering
\includegraphics[width=\linewidth]{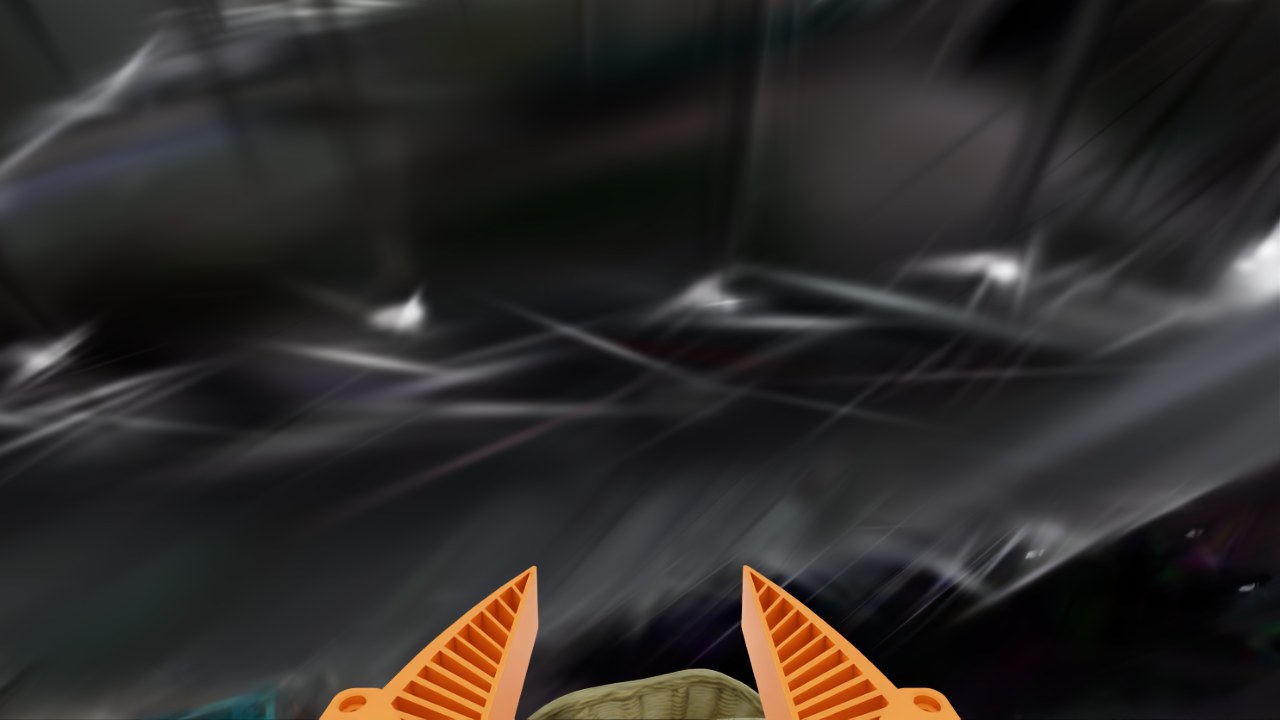}
\par\smallskip\raggedright
\textbf{(a) After gripper opening (step 200).} The gripper is open, but the basket interior is outside the recorded view.
\end{minipage}\hfill
\begin{minipage}[t]{0.485\linewidth}
\vspace{0pt}
\centering
\includegraphics[width=\linewidth]{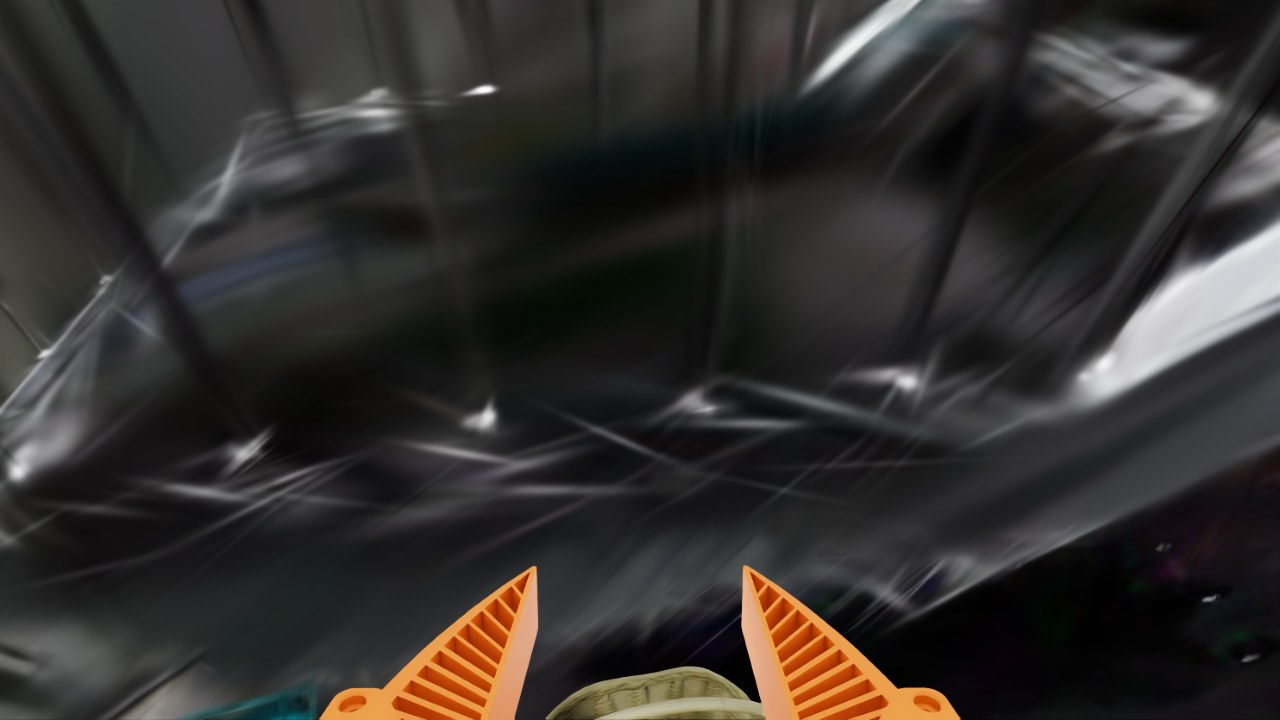}
\par\smallskip\raggedright
\textbf{(b) After upward motion (step 203).} Only part of the basket rim remains visible; the hammer's placement is not shown.
\end{minipage}

\par\medskip\noindent
These selected keyframes show the observation gap, not direct visual evidence of task failure. Failure was established separately by the evaluator. Panel descriptions are author-written explanations, not agent-generated image notes.
\end{caseevidence}

%% file: sections/case_study/k2_case.tex
\paragraph{Acquiring additional outcome observations with AF.}
\label{app:k2_case}
\label{app:k2-evidence}

The task program reported both tool placements as successful. The agent questioned whether the available evidence confirmed either outcome and generated a diagnostic program to inspect the destinations and nearby surfaces. The additional views revealed both tools near floor level, outside their intended destinations, leading the agent to select \texttt{REGENERATE}.

The excerpts below follow this sequence; quotations and code
retain the recorded wording, with omissions marked by
[\ldots] or \texttt{\# [...]}.
Image notes were written by the agent before capture
and express inspection questions, not observed conclusions.

\begin{caseevidence}{1. Task-program output: reported placement success}
\begin{lstlisting}[style=casecode]
hammer done: True
[...]
wrench done: True
\end{lstlisting}
\end{caseevidence}

\begin{caseevidence}{2. Agent's recorded AF analysis: unresolved outcomes}
``The previous program reports both parts done, but neither is confirmed.''

\smallskip
\textbf{Hammer.} [\ldots] ``That shows the hammer left the hand, not that it landed inside the basket. It could have dropped beside the basket or onto the stairs. I need a view looking down into the basket.''

\smallskip
\textbf{Wrench.} [\ldots] ``I need to look at the upper shelf more closely from above, then at the lower shelf and the floor in front of the cabinet, and write down where the wrench actually is.''
\end{caseevidence}

The resulting diagnostic program acquired five views: the upper shelf, lower shelf, cabinet floor, basket from above, and area around the basket. It used map queries to locate the cabinet and basket, then image segmentation and depth to inspect the tools. The helper below connects a requested viewpoint to an agent-written note returned with the image.

\begin{caseevidence}{3. Generated diagnostic code: move, observe, and attach a question}
\begin{lstlisting}[style=casecode,language=Python]
def look(target, offsets, note):
    target = np.asarray(target, float)
    for off in offsets:
        vp = target + np.asarray(off, float)
        if go(vp, target - vp):
            return get_observation(note=note)
    # [...]

fl = np.array([3.55, cc[1], 0.05])
o = look(fl, [[-0.8, 0, 1.0], [-1.0, 0.3, 1.1]],
         "Floor in front of the cabinet's open face: did the wrench fall to the floor?")
report(o, ["wrench"])
\end{lstlisting}
\end{caseevidence}

\noindent Four of the five acquired views are shown below, grouped by task requirement rather than capture order. They show the intended destinations and nearby floor areas, together with the agent-written inspection notes. The lower-shelf view is omitted from the panel. The accompanying measurements provide evidence of the tools' locations.

\casepagebreak

\begin{caseevidence}{4. Returned visual evidence and the agent's image notes}
\begin{minipage}[t]{0.485\linewidth}
\vspace{0pt}
\centering
\includegraphics[width=\linewidth]{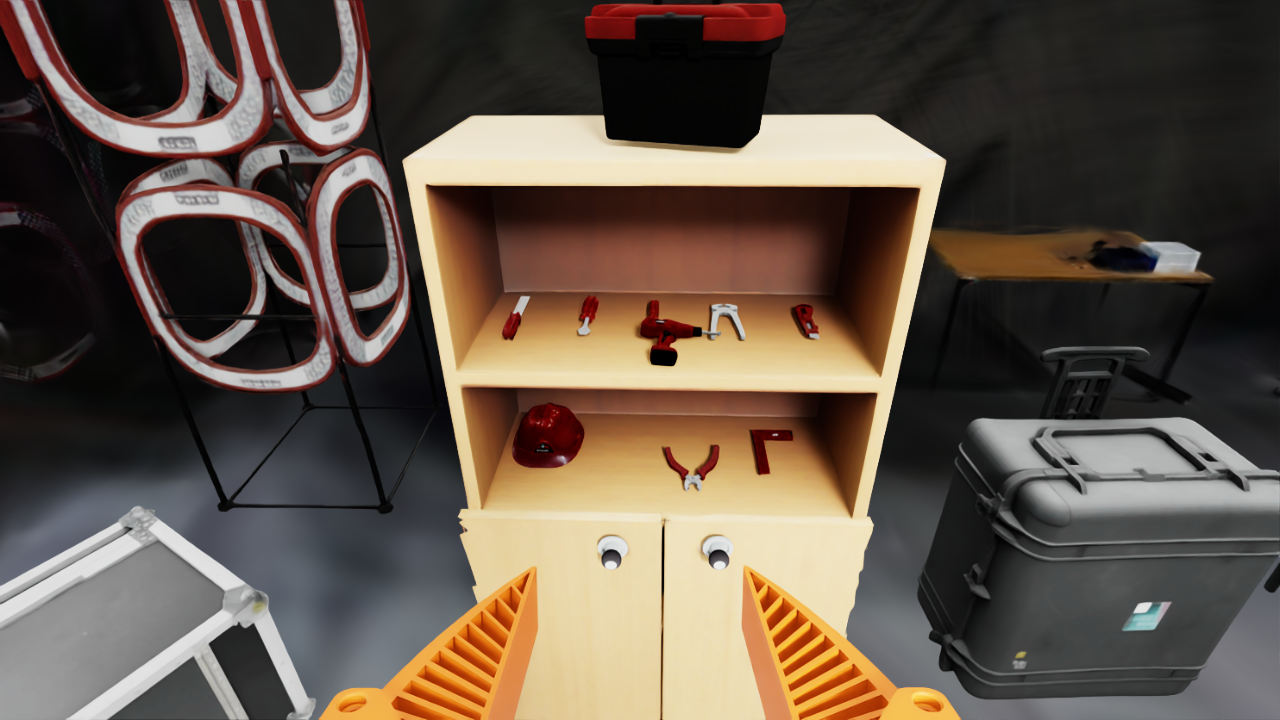}
\par\smallskip\raggedright
\textbf{Upper shelf.} ``Upper open shelf of the cabinet from above-front: is the black-handled wrench from the pegboard on it?''
\end{minipage}\hfill
\begin{minipage}[t]{0.485\linewidth}
\vspace{0pt}
\centering
\includegraphics[width=\linewidth]{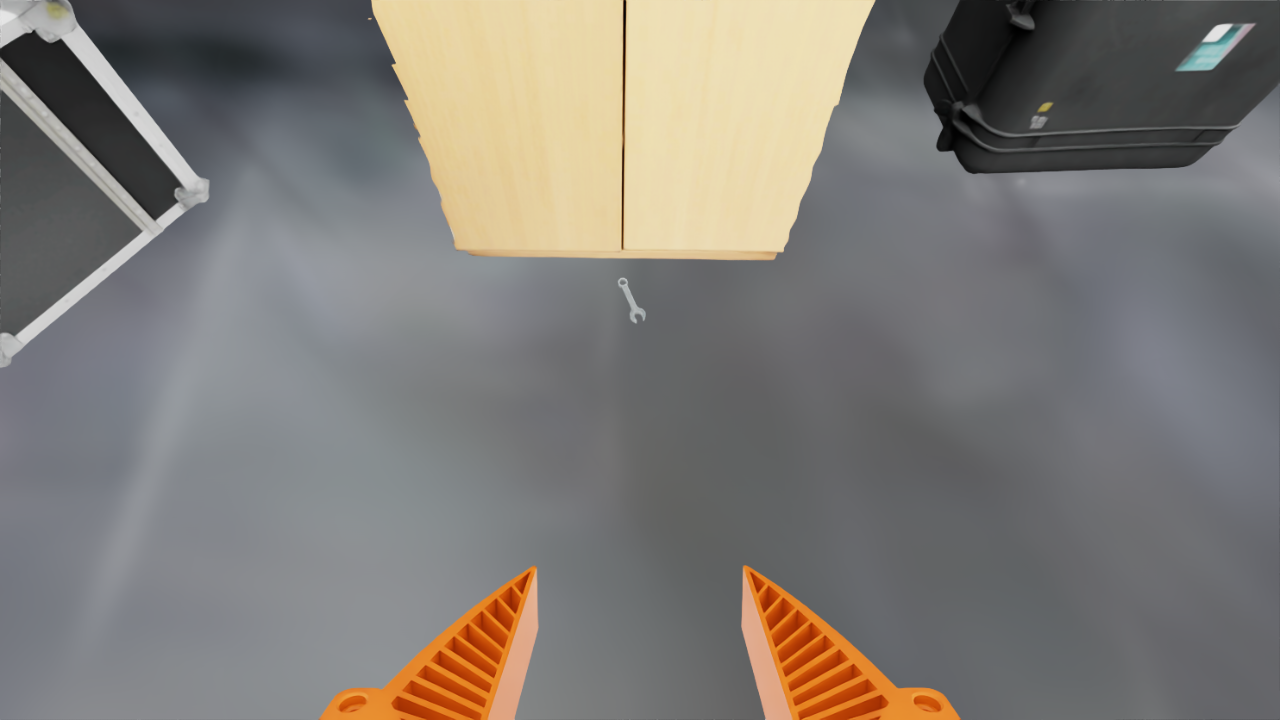}
\par\smallskip\raggedright
\textbf{Cabinet floor.} ``Floor in front of the cabinet's open face: did the wrench fall to the floor?''
\end{minipage}

\par\medskip\noindent
\begin{minipage}[t]{0.485\linewidth}
\vspace{0pt}
\centering
\includegraphics[width=\linewidth]{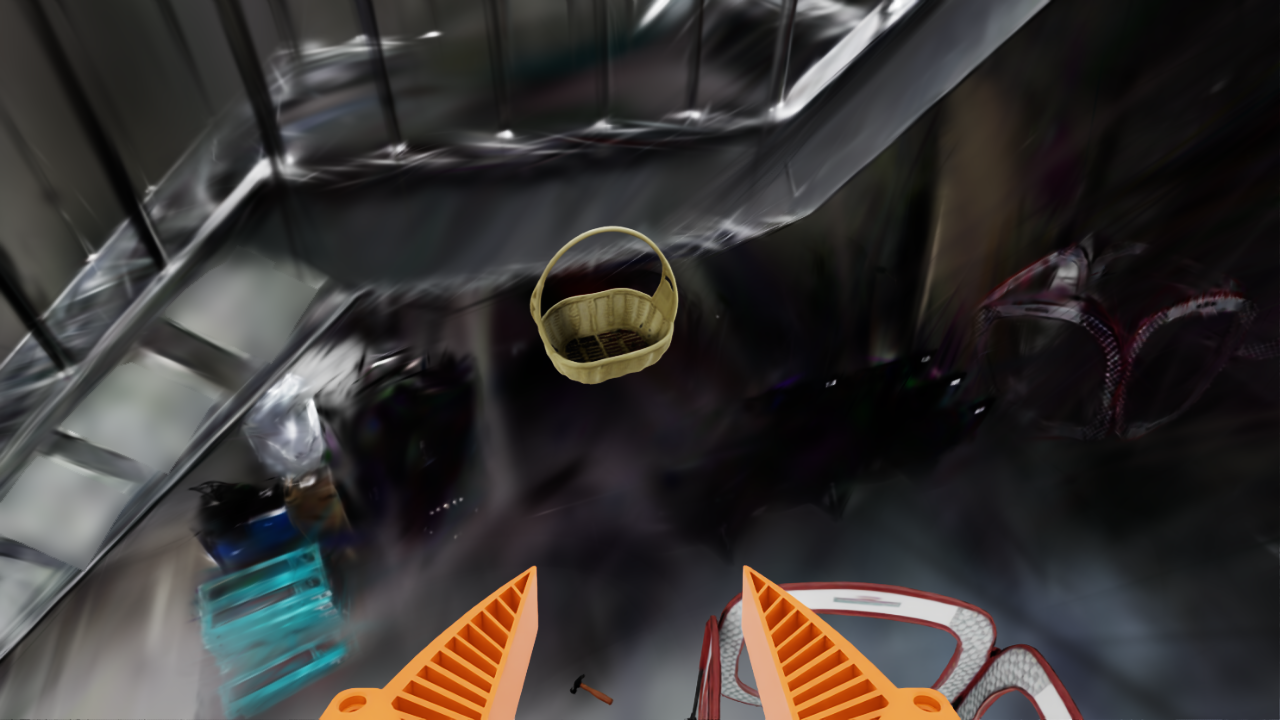}
\par\smallskip\raggedright
\textbf{Basket interior.} ``Basket on the stairs from above: is the hammer inside it?''
\end{minipage}\hfill
\begin{minipage}[t]{0.485\linewidth}
\vspace{0pt}
\centering
\includegraphics[width=\linewidth]{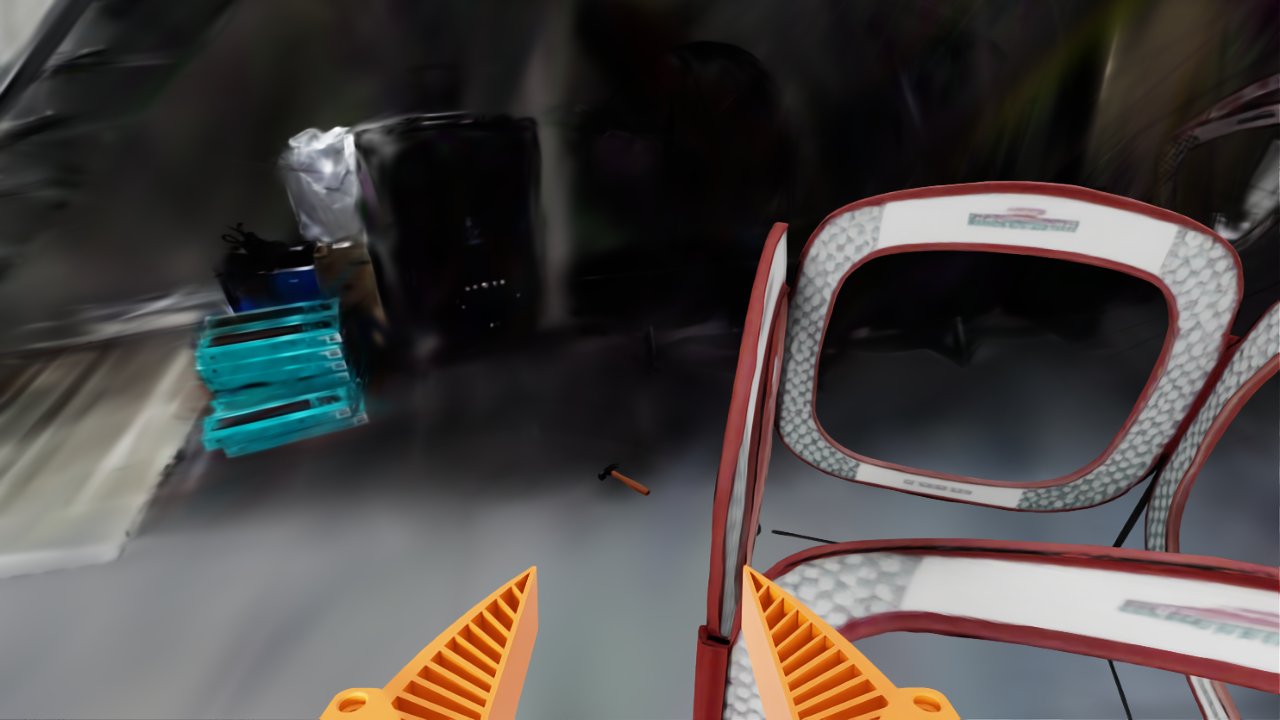}
\par\smallskip\raggedright
\textbf{Area around the basket.} ``Area below/around the basket and stairs: did the hammer fall out instead?''
\end{minipage}
\end{caseevidence}

\begin{caseevidence}{5. Diagnostic console output: estimated tool locations}
\begin{lstlisting}[style=casecode]
[wrench] score 0.91 [...] med [ 3.567  0.953 -0.009] [...]
[...]
[hammer] score 0.92 [...] med [ 2.463  3.979 -0.009] [...]
\end{lstlisting}
\end{caseevidence}

The images and median world-frame coordinates from segmented depth placed both tools near floor level, outside their required destinations. These were agent-visible perception estimates, not ground-truth evaluator outputs. The evidence established unsuccessful placement, without determining when or why either tool fell.

\begin{caseevidence}{6. Decision and an excerpt from the next generated program}
\textbf{Agent decision:} \texttt{REGENERATE}
\begin{lstlisting}[style=casecode,language=Python]
def release_and_hold():
    open_gripper()
    p = pos()
    goto_pose(p, settle=True)          # hold still ~3 s so the object drops free
    get_observation()
    open_gripper()
\end{lstlisting}
\end{caseevidence}

The next program added this release-and-hold routine, revised the hammer release pose, and tried deeper wrench placements. These changes show how the observed failures informed revision; their effectiveness was still to be tested. The immediately following execution failed, and the trial succeeded after further refinement on the fourth task execution. Thus, this case supports AF's role in obtaining evidence against completion and directing revision, while separating a useful diagnosis from an immediately successful repair.

%% file: sections/case_study/align_probe_case.tex
\paragraph{Align: testing alternative contact points.}
\label{app:align-probe-evidence}

After repeated pushes continued to rotate a carton, the agent used AF to test alternative contact locations and sideways slides.
Excerpts retain the recorded wording, with omissions marked
by [\ldots] or \texttt{\# [...]}.

\begin{caseevidence}{1. Agent's recorded AF analysis: which push reduces rotation?}
``My working guess is sideways friction drag.'' [\ldots]

\smallskip
``I need controlled experiments to learn how much each action twists the carton: centre push, high-edge push, low-edge push, and a push followed by a small sideways slide while still in contact.''
\end{caseevidence}

The program tested these actions sequentially, returning to the same viewpoint after each probe to measure the response from RGB-D observations.

\begin{caseevidence}{2. Generated diagnostic code: compare contact actions}
\begin{lstlisting}[style=casecode,language=Python]
open_gripper()
# [...]
exps = [("centre", 0.0, 0.0), ("hi-edge", +1, 0.0),
        ("lo-edge", -1, 0.0), ("centre+drag+y", 0.0, +0.015),
        ("centre+drag-y", 0.0, -0.015), ("hi-edge", +1, 0.0)]
for name, where, drag in exps:
    # [...]
    r = go(G, approach=N, z_approach=0.15, allow_contact=True, settle=True)
    # [...]
    if drag != 0.0 and r:
        G2 = G + np.array([0.0, drag, 0.0])
        go(G2, approach=N, allow_contact=True, settle=True)
        # [...]
    # [...]
    f2, s2 = measure_B("after " + name)
\end{lstlisting}
\end{caseevidence}

The first high-$y$ edge push produced less rotation per unit of advance than the centre and low-$y$ edge pushes. Neither sideways slide reversed the rotation.

\begin{caseevidence}{3. Returned diagnostic views and the agent's image notes}
\begin{minipage}[t]{0.485\linewidth}
\vspace{0pt}\centering
\includegraphics[width=\linewidth]{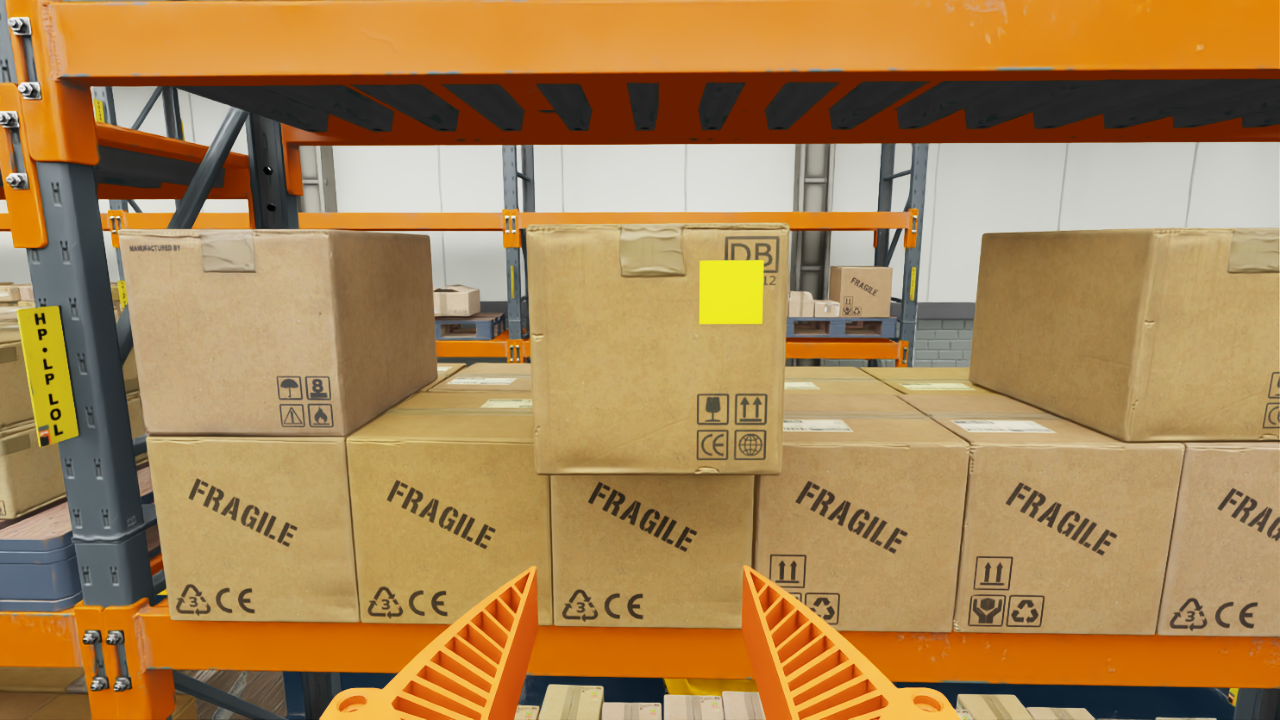}
\par\smallskip\raggedright
\textbf{Before probing.} ``carton B untouched, head-on, before the push experiments''
\end{minipage}\hfill
\begin{minipage}[t]{0.485\linewidth}
\vspace{0pt}\centering
\includegraphics[width=\linewidth]{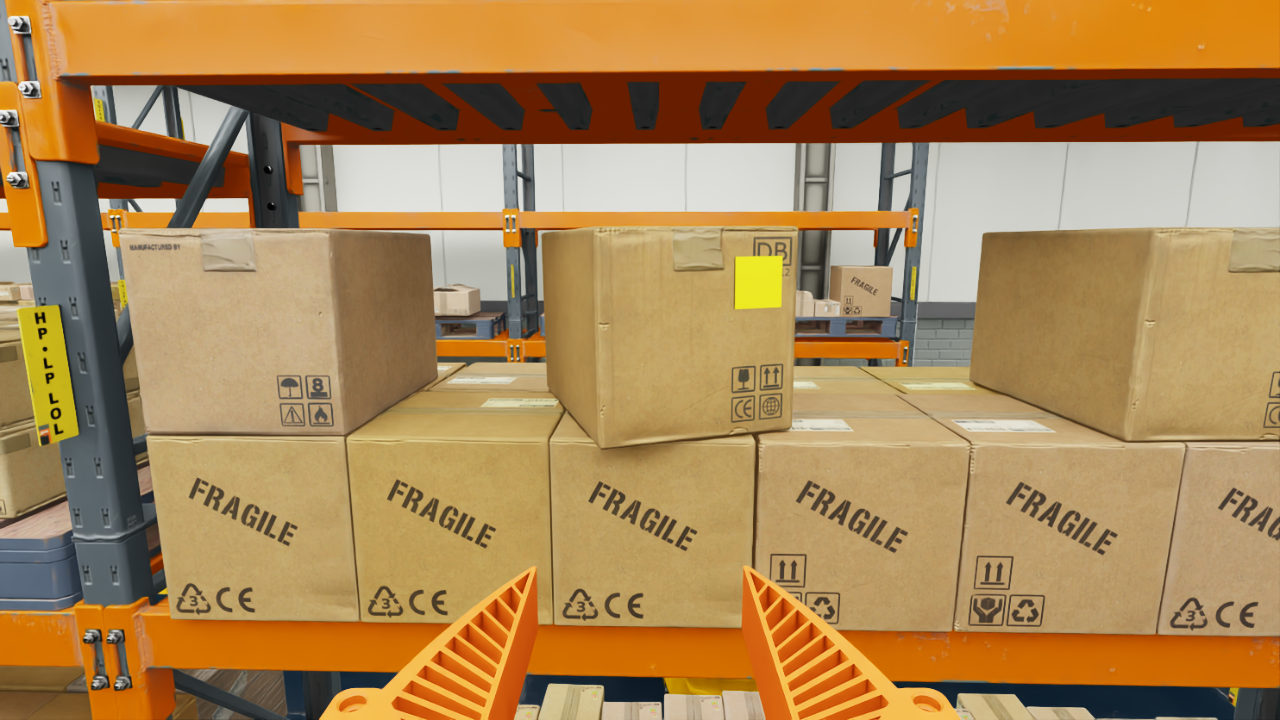}
\par\smallskip\raggedright
\textbf{After probing.} ``carton B after the six push experiments (centre, hi-edge, lo-edge, drag +y, drag -y, hi-edge)''
\end{minipage}
\end{caseevidence}

\begin{caseevidence}{4. Decision and subsequent program revision}
\textbf{Agent decision:} \texttt{REGENERATE}

\smallskip
The next program initialized a linear push-response model
from the diagnostic observations and used it to select
the lateral contact position.

\begin{lstlisting}[style=casecode,language=Python]
twist_data = [(0.0, -1.0, 1.0), (0.8, -0.2, 1.0), (-0.8, -1.7, 0.5)]
def twist_model():
    # [...]
    beta, alpha = np.polyfit(u, r, 1, w=np.sqrt(w))
    # [...]

# [...]
        w = y2 - y1; yc = 0.5 * (y1 + y2)
        # [...]
        alpha, beta = twist_model()
        # [...]
        u = float(np.clip((rate_des - alpha) / beta, -0.82, 0.82))
        yp = yc + u * w / 2.0
\end{lstlisting}

In the next task execution, the robot pushed on the high-$y$
side, toward which the carton's center of mass was shifted,
and precisely aligned the boxes.
\end{caseevidence}

%% file: sections/case_study/payload_case.tex
\paragraph{Payload: testing a load capacity.}
\label{app:payload_case}
\label{app:payload-evidence}

The agent used AF to test whether the robot could lift
a proposed basket load. The result informed a revised transport plan;
a later AF round checked the final placements before the agent
declared completion.
Excerpts retain the recorded wording, with omissions marked
by [\ldots] or \texttt{\# [...]}.

\begin{caseevidence}{1. Agent's recorded AF analysis: a plan depends on an unknown limit}
``The packing I planned needs baskets of 1.2 kg (red + yellow) and 1.0 kg (blue + black) with green carried alone. That only works if the limit is at least 1.2 kg.'' [\ldots]

\smallskip
``I'll load red and yellow into the $-y$ basket and test-lift it ($\approx$1.2 kg).'' [\ldots]
\end{caseevidence}

The preceding execution had measured the bottle masses but stopped at an array-comparison error. Beyond fixing that error, the agent needed to check its proposed packing. It generated code to load the red and yellow bottles, grasp the basket handle, attempt a lift, and return the interface's response with a diagnostic image.

\begin{caseevidence}{2. Generated diagnostic code: test the load and report the response}
\begin{lstlisting}[style=casecode,language=Python]
r, txt = captured(lambda: go(ap + np.array([0, 0, 0.2]), settle=True))
lim = "limit exceeded" in txt
P(f"LIFT TEST {tag}: result={r} kg={parse_kg(txt)} limit_exceeded={lim}")
get_observation(note=f"lift test {tag}: basket lifted? limit_exceeded={lim}")
\end{lstlisting}
\end{caseevidence}

\begin{caseevidence}{3. Diagnostic response and the agent's image note}
\begin{lstlisting}[style=casecode]
[drone thrust sensor] limit exceeded: the drone cannot lift what it is holding. Nothing moved.
LIFT TEST basket+red+yellow ~1.2kg: result=refused kg=None limit_exceeded=True
\end{lstlisting}
\smallskip
\begin{minipage}[c]{0.43\linewidth}
\includegraphics[width=\linewidth]{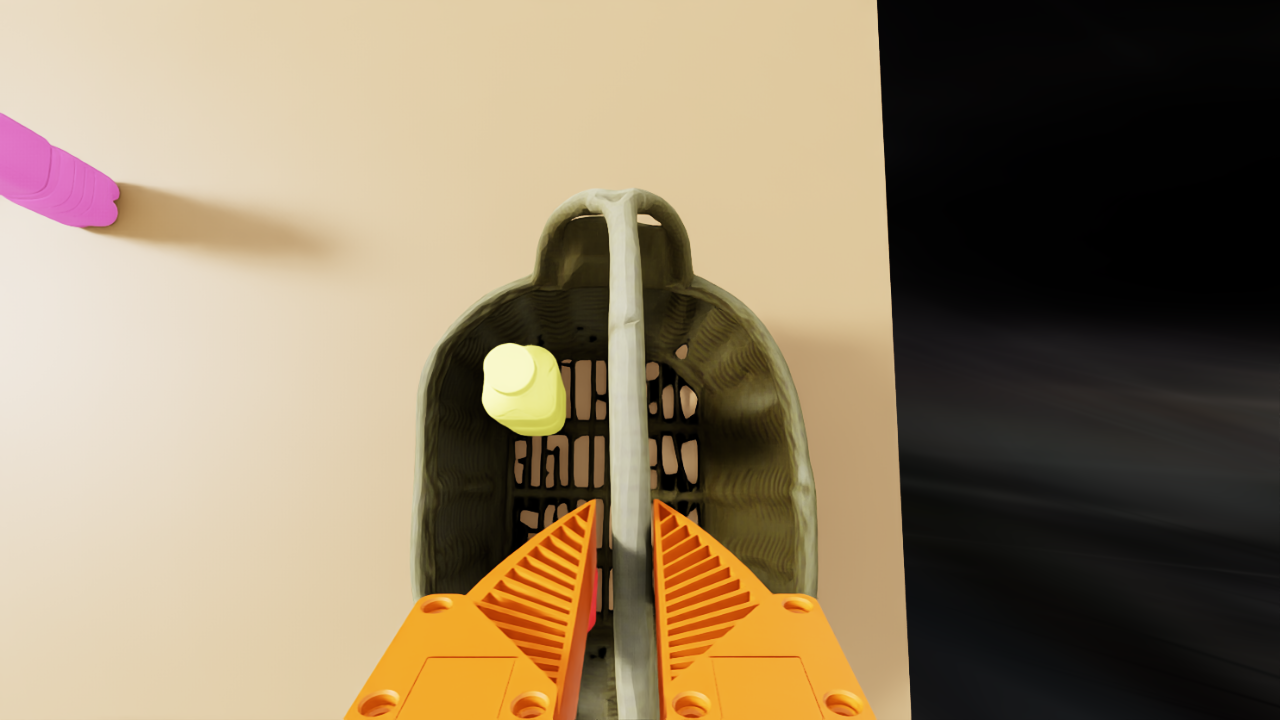}
\end{minipage}\hfill
\begin{minipage}[c]{0.53\linewidth}
\textbf{Agent-written note:}\par
``lift test basket+red+yellow $\sim$1.2kg: basket lifted? limit\_exceeded=True''
\end{minipage}
\end{caseevidence}

The refusal showed that the proposed load was infeasible; it did not identify the exact carrying limit. A subsequent intended 1.5\,kg test failed to grasp the handle, so it supplied no additional capacity bound. The agent selected \texttt{REGENERATE}.

\casepagebreak

\begin{caseevidence}{4. Next generated program: revise the planning limit}
\begin{lstlisting}[style=casecode,language=Python]
# payload: 0.9 kg bottle alone lifts, basket+1.1 kg of bottles (1.2 kg) is refused -> start from 1.1 kg,
# and lower it automatically if the thrust sensor refuses a lift.
PAYLOAD_LIMIT = [1.1]
BASKET_KG = 0.1
# [...]
best = plan(pool, R, PAYLOAD_LIMIT[0])
\end{lstlisting}
\end{caseevidence}

The next program replaced its previous 1.5\,kg assumption with a conservative 1.1\,kg planning limit. It paired red with black and blue with yellow, and carried green separately. Both basket lifts succeeded, as recorded below. The revision also fixed the array-comparison error.

\begin{caseevidence}{5. Next task execution: both planned basket loads lift successfully}
\begin{lstlisting}[style=casecode]
basket0 lift: reached kg=1.1 limit_exceeded=False load~1.10
[...]
basket1 lift: reached kg=1.1 limit_exceeded=False load~1.10
\end{lstlisting}
\end{caseevidence}

\begin{caseevidence}{6. Next AF round: verify the distributed completion conditions}
``The run finished without errors, but I haven't confirmed the final state yet.'' [\ldots]

\smallskip
``I haven't seen both baskets full on the white table in one view.'' [\ldots] ``I haven't checked that only pink, purple and white remain there.'' [\ldots]

\smallskip
\begin{minipage}[t]{0.485\linewidth}
\includegraphics[width=\linewidth]{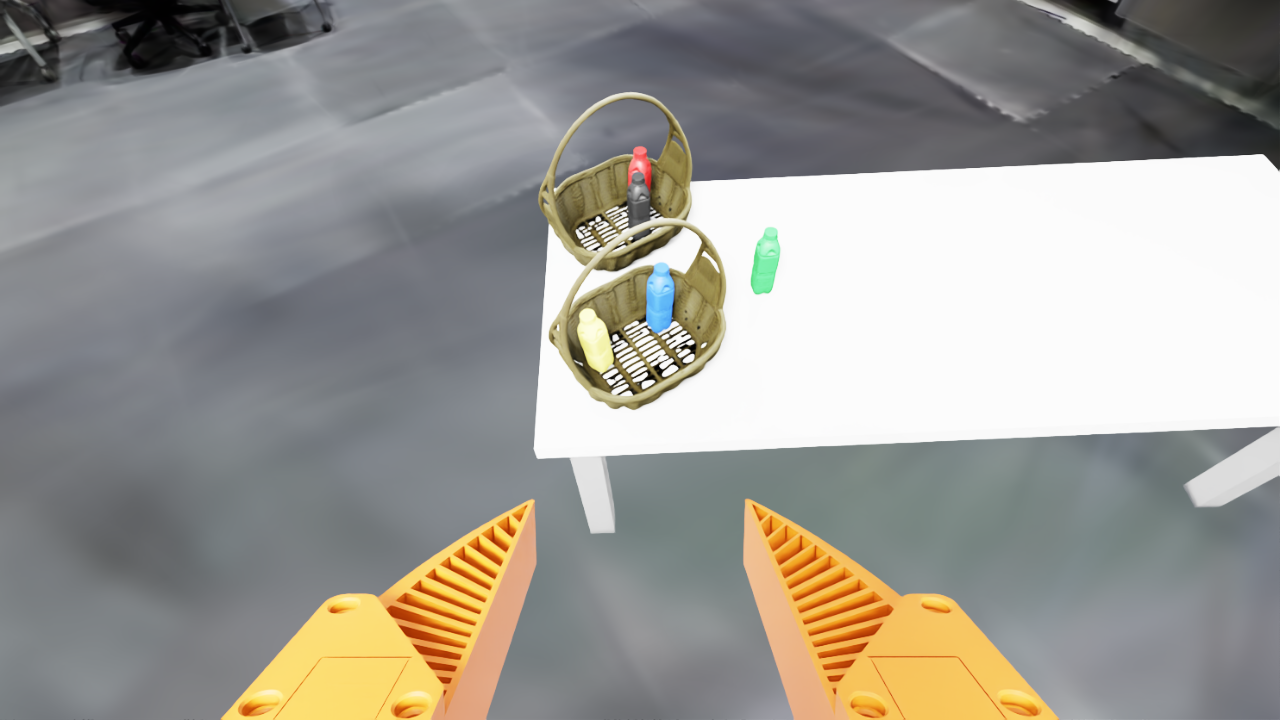}
\par\smallskip
\textbf{Destination note:} ``white table overview: both baskets (red+black, blue+yellow) and green bottle present?''
\end{minipage}\hfill
\begin{minipage}[t]{0.485\linewidth}
\includegraphics[width=\linewidth]{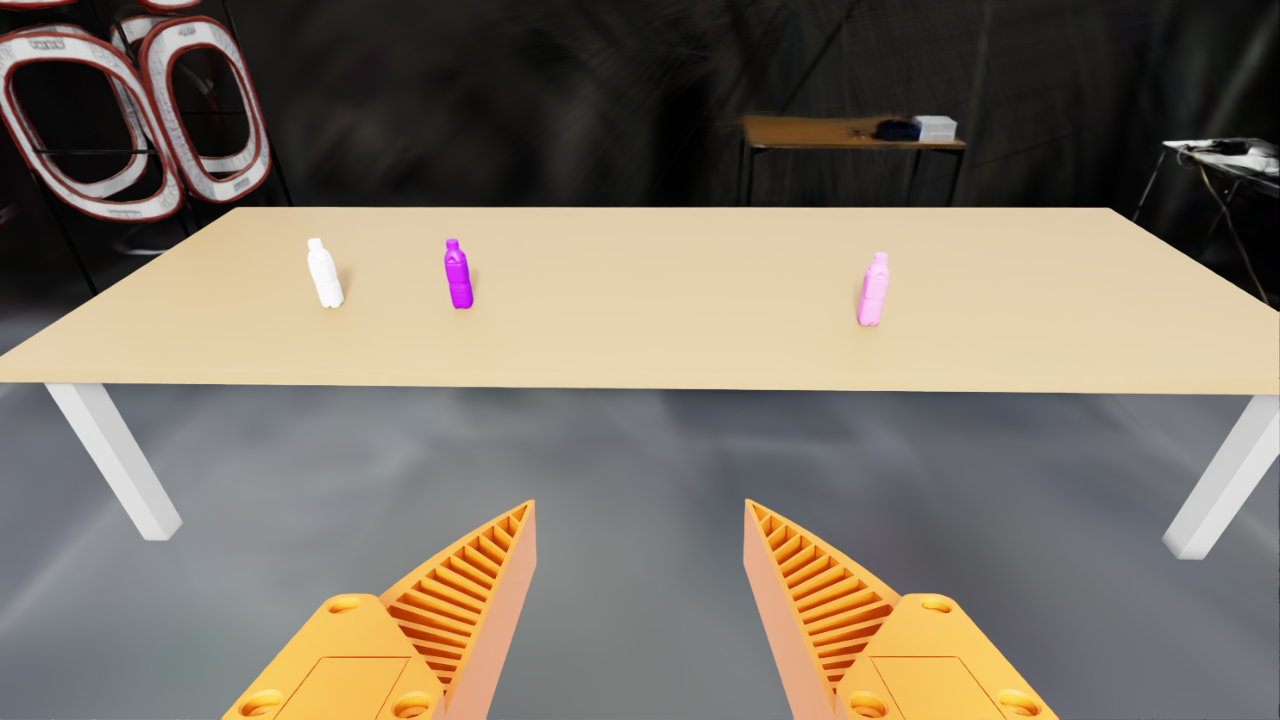}
\par\smallskip
\textbf{Source note:} ``brown table: only the 3 empty bottles (pink, purple, white) should remain''
\end{minipage}
\end{caseevidence}

AF obtained two destination views and one source view, with image-based detections supporting five bottles and two baskets at the destination and three empty bottles at the source. The agent then declared \texttt{FINISH}; the supplied evaluation report records this execution as successful. The sequence illustrates how AF can first test a planning constraint and later check completion, rather than treating a successful lift or an error-free program return as sufficient evidence for the full task.

%% file: references.bib
@misc{anthropic2026opus5,
  author = {{Anthropic}},
  title = {Introducing {Claude Opus 5}},
  year = {2026},
  url = {https://www.anthropic.com/news/claude-opus-5},
  note = {Accessed September 26, 2026},
}

@misc{anthropic2026opus55,
  author = {{Anthropic}},
  title = {Introducing {Claude Opus 5.5}},
  year = {2026},
  url = {https://www.anthropic.com/claude-opus-5-5},
  note = {Accessed September 26, 2026},
}

@misc{google2026gemini38flash,
  author = {{Google DeepMind}},
  title = {{Gemini 3.8 Flash} Model Card},
  year = {2026},
  url = {https://deepmind.google/models/model-cards/gemini-3-8-flash/},
  note = {Accessed September 26, 2026},
}

@article{bajcsy1988active,
  author = {Bajcsy, Ruzena},
  title = {Active perception},
  journal = {Proceedings of the IEEE},
  volume = {76},
  number = {8},
  pages = {996--1005},
  year = {1988},
}

@article{bajcsy2018revisiting,
  author = {Bajcsy, Ruzena and Aloimonos, Yiannis and Tsotsos, John K.},
  title = {Revisiting active perception},
  journal = {Autonomous Robots},
  volume = {42},
  number = {2},
  pages = {177--196},
  year = {2018},
}

@inproceedings{bauer2025opensource,
  author = {Bauer, Erik and Bl{\"o}chlinger, Marc and Strauch, Pascal and Raayatsanati, Arman and Cavelti, Curdin and Katzschmann, Robert K.},
  title = {An open-source soft robotic platform for autonomous aerial manipulation in the wild},
  booktitle = {Conference on Robot Learning (CoRL)},
  series = {Proceedings of Machine Learning Research},
  volume = {270},
  pages = {3094--3106},
  year = {2025},
}

@article{bodie2021active,
  author = {Bodie, Karen and Brunner, Maximilian and Pantic, Michael and Walser, Stefan and Pf{\"a}ndler, Patrick and Angst, Ueli and Siegwart, Roland and Nieto, Juan},
  title = {Active interaction force control for contact-based inspection with a fully actuated aerial vehicle},
  journal = {IEEE Transactions on Robotics},
  volume = {37},
  number = {3},
  pages = {709--722},
  year = {2021},
}

@article{bohg2017interactive,
  author = {Bohg, Jeannette and Hausman, Karol and Sankaran, Bharath and Brock, Oliver and Kragic, Danica and Schaal, Stefan and Sukhatme, Gaurav S.},
  title = {Interactive perception: Leveraging action in perception and perception in action},
  journal = {IEEE Transactions on Robotics},
  volume = {33},
  number = {6},
  pages = {1273--1291},
  year = {2017},
}

@misc{chen2024roboscript,
  author = {Chen, Junting and Mu, Yao and Yu, Qiaojun and Wei, Tianming and Wu, Silang and Yuan, Zhecheng and Liang, Zhixuan and Yang, Chao and Zhang, Kaipeng and Shao, Wenqi and others},
  title = {{RoboScript}: Code generation for free-form manipulation tasks across real and simulation},
  year = {2024},
  howpublished = {arXiv preprint arXiv:2402.14623},
}

@misc{chen2026few,
  author = {Chen, Daphne and Jain, Archit Ritesh and Goossen, Eric and Romig, Emma and Murray, Michael and Walker, Nick and Cakmak, Maya},
  title = {A few words go a long way: Language guided robot policy synthesis},
  year = {2026},
  howpublished = {arXiv preprint arXiv:2607.23784},
}

@article{deng2025whole,
  author = {Deng, Weiliang and Chen, Hongming and Ye, Biyu and Chen, Haoran and Li, Ziliang and Lyu, Ximin},
  title = {Whole-body integrated motion planning for aerial manipulators},
  journal = {IEEE Transactions on Robotics},
  volume = {41},
  pages = {6661--6679},
  year = {2025},
}

@inproceedings{detone2026boxer,
  author = {DeTone, Daniel and Shen, Tianwei and Zhang, Fan and Ma, Lingni and Straub, Julian and Newcombe, Richard and Engel, Jakob},
  title = {{Boxer}: Robust lifting of open-world {2D} bounding boxes to {3D}},
  booktitle = {European Conference on Computer Vision (ECCV)},
  year = {2026},
  note = {Accepted; proceedings forthcoming},
}

@inproceedings{fu2026capx,
  author = {Fu, Letian and Yu, Justin and El-Refai, Karim and Kou, Ethan and Xue, Haoru and Huang, Huang and Xiao, Wenli and Wang, Guanzhi and Niu, Dantong and Li, Fei-Fei and others},
  title = {{CaP-X}: A framework for benchmarking and improving coding agents for robot manipulation},
  booktitle = {International Conference on Machine Learning (ICML)},
  year = {2026},
}

@inproceedings{goodarzi2013geometric,
  author = {Goodarzi, Farhad and Lee, Daewon and Lee, Taeyoung},
  title = {Geometric nonlinear {PID} control of a quadrotor {UAV} on {SE(3)}},
  booktitle = {European Control Conference (ECC)},
  pages = {3845--3850},
  year = {2013},
}

@article{guo2024flying,
  author = {Guo, Xiaofeng and He, Guanqi and Xu, Jiahe and Mousaei, Mohammadreza and Geng, Junyi and Scherer, Sebastian and Shi, Guanya},
  title = {Flying calligrapher: Contact-aware motion and force planning and control for aerial manipulation},
  journal = {IEEE Robotics and Automation Letters},
  volume = {9},
  number = {12},
  pages = {11194--11201},
  year = {2024},
}

@inproceedings{gupta2026umi,
  author = {Gupta, Harsh and Guo, Xiaofeng and Ha, Huy and Pan, Chuer and Cao, Muqing and Lee, Dongjae and Scherer, Sebastian and Song, Shuran and Shi, Guanya},
  title = {{UMI-on-Air}: Embodiment-aware guidance for embodiment-agnostic visuomotor policies},
  booktitle = {IEEE International Conference on Robotics and Automation (ICRA)},
  year = {2026},
}

@article{hart1968formal,
  author = {Hart, Peter E. and Nilsson, Nils J. and Raphael, Bertram},
  title = {A formal basis for the heuristic determination of minimum cost paths},
  journal = {IEEE Transactions on Systems Science and Cybernetics},
  volume = {4},
  number = {2},
  pages = {100--107},
  year = {1968},
}

@inproceedings{he2025flying,
  author = {He, Guanqi and Guo, Xiaofeng and Tang, Luyi and Zhang, Yuanhang and Mousaei, Mohammadreza and Xu, Jiahe and Geng, Junyi and Scherer, Sebastian and Shi, Guanya},
  title = {Flying hand: End-effector-centric framework for versatile aerial manipulation teleoperation and policy learning},
  booktitle = {Robotics: Science and Systems (RSS)},
  year = {2025},
}

@article{hu2024metric3dv2,
  author = {Hu, Mu and Yin, Wei and Zhang, Chi and Cai, Zhipeng and Long, Xiaoxiao and Chen, Hao and Wang, Kaixuan and Yu, Gang and Shen, Chunhua and Shen, Shaojie},
  title = {{Metric3D v2}: A versatile monocular geometric foundation model for zero-shot metric depth and surface normal estimation},
  journal = {IEEE Transactions on Pattern Analysis and Machine Intelligence},
  volume = {46},
  number = {12},
  pages = {10579--10596},
  year = {2024},
}

@article{hwang2025autonomous,
  author = {Hwang, Sunwoo and Lee, Dongjae and Kim, Changhyeon and Kim, Hyoun Jin},
  title = {Autonomous heavy object pushing using a coaxial tiltrotor},
  journal = {IEEE Transactions on Automation Science and Engineering},
  volume = {22},
  pages = {4243--4256},
  year = {2025},
}

@inproceedings{izquierdo2025mvsanywhere,
  author = {Izquierdo, Sergio and Sayed, Mohamed and Firman, Michael and Garcia-Hernando, Guillermo and Turmukhambetov, Daniyar and Civera, Javier and Mac Aodha, Oisin and Brostow, Gabriel and Watson, Jamie},
  title = {{MVSAnywhere}: Zero-shot multi-view stereo},
  booktitle = {IEEE/CVF Conference on Computer Vision and Pattern Recognition (CVPR)},
  pages = {11493--11504},
  year = {2025},
}

@misc{jax2018github,
  author = {Bradbury, James and Frostig, Roy and Hawkins, Peter and Johnson, Matthew James and Katariya, Yash and Leary, Chris and Maclaurin, Dougal and Necula, George and Paszke, Adam and Vander{P}las, Jake and others},
  title = {{JAX}: Composable transformations of {Python}+{NumPy} programs},
  year = {2018},
  howpublished = {Software},
  url = {https://github.com/jax-ml/jax},
}

@inproceedings{kim2025pyroki,
  author = {Kim, Chung Min and Yi, Brent and Choi, Hongsuk and Ma, Yi and Goldberg, Ken and Kanazawa, Angjoo},
  title = {{PyRoki}: A modular toolkit for robot kinematic optimization},
  booktitle = {IEEE/RSJ International Conference on Intelligent Robots and Systems (IROS)},
  pages = {1312--1319},
  year = {2025},
}

@article{lee2026autonomous,
  author = {Lee, Dongjae and Kim, Byeongjun and Kim, Hyoun Jin},
  title = {Autonomous aerial manipulation at arbitrary pose in {SE(3)} with robust control and whole-body planning},
  journal = {The International Journal of Robotics Research},
  volume = {45},
  number = {7},
  pages = {1021--1045},
  year = {2026},
}

@inproceedings{liang2023code,
  author = {Liang, Jacky and Huang, Wenlong and Xia, Fei and Xu, Peng and Hausman, Karol and Ichter, Brian and Florence, Pete and Zeng, Andy},
  title = {Code as policies: Language model programs for embodied control},
  booktitle = {IEEE International Conference on Robotics and Automation (ICRA)},
  pages = {9493--9500},
  year = {2023},
}

@misc{lin2026physcap,
  author = {Lin, Chen-Yu and Chen, Jing-Wen and Chang, Hsueh-En and Chen, Hung-An and Chang, Sheng-Hsun and Huang, Chi-Pin and Yang, Fu-En and Chen, Min-Hung and Chen, Yi-Ting and Wang, Yu-Chiang Frank and others},
  title = {{PhysCaP}: Grounding code-as-policy agent with physics-informed exploration},
  year = {2026},
  howpublished = {arXiv preprint arXiv:2608.21031},
}

@inproceedings{liu2023reflect,
  author = {Liu, Zeyi and Bahety, Arpit and Song, Shuran},
  title = {{REFLECT}: Summarizing robot experiences for failure explanation and correction},
  booktitle = {Conference on Robot Learning (CoRL)},
  series = {Proceedings of Machine Learning Research},
  volume = {229},
  pages = {3468--3484},
  year = {2023},
}

@inproceedings{liu2026activevla,
  author = {Liu, Zhenyang and Gu, Yongchong and Wang, Yikai and Xue, Xiangyang and Fu, Yanwei},
  title = {{ActiveVLA}: Injecting active perception into vision-language-action models for precise {3D} robotic manipulation},
  booktitle = {IEEE/CVF Conference on Computer Vision and Pattern Recognition (CVPR)},
  pages = {8141--8151},
  year = {2026},
}

@inproceedings{liu2026sapave,
  author = {Liu, Mengzhen and Zhou, Enshen and Chi, Cheng and Han, Yi and Rong, Shanyu and Chen, Liming and Wang, Pengwei and Wang, Zhongyuan and Zhang, Shanghang},
  title = {{SaPaVe}: Towards active perception and manipulation in vision-language action models for robotics},
  booktitle = {IEEE/CVF Conference on Computer Vision and Pattern Recognition (CVPR)},
  pages = {37164--37174},
  year = {2026},
}

@misc{lu2026aspire,
  author = {Lu, Runyu and Wu, Yubo and Kou, Ethan and Fu, Letian and Xiao, Wenli and Mandlekar, Ajay and Xu, Yinzhen and Shi, Guanya and Goldberg, Ken and Chen, Ang and others},
  title = {{ASPIRE}: Agentic {/Skills} discovery for robotics},
  year = {2026},
  howpublished = {arXiv preprint arXiv:2607.00272},
}

@inproceedings{minderer2024OWLv2,
  author = {Minderer, Matthias and Gritsenko, Alexey and Houlsby, Neil},
  title = {Scaling open-vocabulary object detection},
  booktitle = {Advances in Neural Information Processing Systems},
  volume = {36},
  pages = {72983--73007},
  year = {2023},
}

@misc{mishra2025aermani,
  author = {Mishra, Sarthak and Yadav, Rishabh Dev and Das, Avirup and Gupta, Saksham and Pan, Wei and Roy, Spandan},
  title = {{AERMANI-VLM}: Structured prompting and reasoning for aerial manipulation with vision language models},
  year = {2025},
  howpublished = {arXiv preprint arXiv:2511.01472},
}

@article{mower2026rosllm,
  author = {Mower, Christopher E. and Wan, Yuhui and Yu, Hongzhan and Grosnit, Antoine and Gonzalez-Billandon, Jonas and Zimmer, Matthieu and Liu, Puze and Palenicek, Daniel and Tateo, Davide and Peters, Jan and others},
  title = {A robot operating system framework for using large language models in embodied {AI}},
  journal = {Nature Machine Intelligence},
  volume = {8},
  number = {3},
  pages = {313--325},
  year = {2026},
}

@inproceedings{mu2024robocodex,
  author = {Mu, Yao and Chen, Junting and Zhang, Qing-Long and Chen, Shoufa and Yu, Qiaojun and Ge, Chongjian and Chen, Runjian and Liang, Zhixuan and Hu, Mengkang and Tao, Chaofan and others},
  title = {{RoboCodeX}: Multimodal code generation for robotic behavior synthesis},
  booktitle = {International Conference on Machine Learning (ICML)},
  series = {Proceedings of Machine Learning Research},
  volume = {235},
  pages = {36434--36454},
  year = {2024},
}

@misc{nvidia2025isaaclabgpuacceleratedsimulation,
  author = {Mittal, Mayank and Roth, Pascal and Tigue, James and Richard, Antoine and Zhang, Octi and Du, Peter and Serrano-Mu\~noz, Antonio and Yao, Xinjie and Zurbr\"ugg, Ren\'e and Rudin, Nikita and others},
  title = {{Isaac Lab}: A {GPU}-accelerated simulation framework for multi-modal robot learning},
  year = {2025},
  howpublished = {arXiv preprint arXiv:2511.04831},
}

@misc{nvidia_isaac_sim_assets,
  author = {{NVIDIA}},
  title = {{Isaac Sim} 6.0.0 documentation: {Isaac Sim} assets},
  year = {2026},
  url = {https://docs.isaacsim.omniverse.nvidia.com/6.0.0/assets/usd_assets_overview.html},
  note = {Accessed September 26, 2026},
}

@misc{nvidia_isaac_sim_warehouse,
  author = {{NVIDIA}},
  title = {{Isaac Sim} 5.1.0 documentation: Environment assets},
  year = {2026},
  url = {https://docs.isaacsim.omniverse.nvidia.com/5.1.0/assets/usd_assets_environments.html},
  note = {Accessed September 26, 2026},
}

@inproceedings{oelerich2025languageguided,
  author = {Oelerich, Thies and Hartl-Nesic, Christian and Kugi, Andreas},
  title = {Language-guided manipulator motion planning with bounded task space},
  booktitle = {Conference on Robot Learning (CoRL)},
  series = {Proceedings of Machine Learning Research},
  volume = {270},
  pages = {749--779},
  year = {2025},
}

@article{ollero2022past,
  author = {Ollero, Anibal and Tognon, Marco and Suarez, Alejandro and Lee, Dongjun and Franchi, Antonio},
  title = {Past, present, and future of aerial robotic manipulators},
  journal = {IEEE Transactions on Robotics},
  volume = {38},
  number = {1},
  pages = {626--645},
  year = {2022},
}

@misc{openai2026astra,
  author = {{OpenAI}},
  title = {{GPT-6 Astra}: A new generation of intelligence},
  year = {2026},
  url = {https://openai.com/index/gpt-6-astra/},
  note = {Accessed September 26, 2026},
}

@inproceedings{radl2026come,
  author = {Radl, Lukas and Windisch, Felix and Kurz, Andreas and K{\"o}hler, Thomas and Steiner, Michael and Steinberger, Markus},
  title = {Confidence-based mesh extraction from {3D} {Gaussians}},
  booktitle = {European Conference on Computer Vision (ECCV)},
  series = {Lecture Notes in Computer Science},
  volume = {17039},
  pages = {307--325},
  year = {2026},
}

@inproceedings{sam3dteam2025sam3d3dfyimages,
  author = {Chen, Xingyu and Chu, Fu-Jen and Gleize, Pierre and Liang, Kevin J and Sax, Alexander and Tang, Hao and Wang, Weiyao and Guo, Michelle and Hardin, Thibaut and Li, Xiang and others},
  title = {{SAM 3D}: {3Dfy} anything in images},
  booktitle = {IEEE/CVF Conference on Computer Vision and Pattern Recognition (CVPR)},
  pages = {7220--7232},
  year = {2026},
}

@misc{schlesinger2024roborepair,
  author = {Schlesinger, Claire and Guha, Arjun and Biswas, Joydeep},
  title = {Creating and repairing robot programs in open-world domains},
  year = {2024},
  howpublished = {arXiv preprint arXiv:2410.18893},
}

@inproceedings{singh2023progprompt,
  author = {Singh, Ishika and Blukis, Valts and Mousavian, Arsalan and Goyal, Ankit and Xu, Danfei and Tremblay, Jonathan and Fox, Dieter and Thomason, Jesse and Garg, Animesh},
  title = {{ProgPrompt}: Generating situated robot task plans using large language models},
  booktitle = {IEEE International Conference on Robotics and Automation (ICRA)},
  pages = {11523--11530},
  year = {2023},
}

@misc{singh2026aerograb,
  author = {Singh, Shivansh Pratap and Nair, Naveen Sudheer and Ujjawal, Samaksh and Mishra, Sarthak and Patil, Soham and Yadav, Rishabh Dev and Roy, Spandan},
  title = {{AeroGrab}: A unified framework for aerial grasping in cluttered environments},
  year = {2026},
  howpublished = {arXiv preprint arXiv:2603.15097},
}

@misc{tucker2026pimakefly,
  author = {Tucker, Johnathan and Liu, Denis and Swann, Aiden and Ren, Allen and Yu, Javier and Sun, Jiankai and Kim, Brandon and McGranahan, Lachlain and Vuong, Quan and Schwager, Mac},
  title = {{$\pi$}, but make it fly: Physics-guided transfer of {VLA} models to aerial manipulation},
  year = {2026},
  howpublished = {arXiv preprint arXiv:2603.25038},
}

@article{ubellacker2024highspeed,
  author = {Ubellacker, Samuel and Ray, Aaron and Bern, James M. and Strader, Jared and Carlone, Luca},
  title = {High-speed aerial grasping using a soft drone with onboard perception},
  journal = {npj Robotics},
  volume = {2},
  year = {2024},
  note = {Article 5},
}

@article{wang2025observe,
  author = {Wang, Guokang and Li, Hang and Zhang, Shuyuan and Guo, Di and Liu, Yanhong and Liu, Huaping},
  title = {Observe then act: Asynchronous active vision-action model for robotic manipulation},
  journal = {IEEE Robotics and Automation Letters},
  volume = {10},
  number = {4},
  pages = {3422--3429},
  year = {2025},
}

@inproceedings{wang2026ambench,
  author = {Wang, Yutong and Lee, Dongjae and Guo, Xiaofeng and Zhan, Yuanzhu and Jiang, Yufei and Saravanan, Bavin and Cao, Muqing and Xie, Jia and Mao, Chenyang and Scherer, Sebastian and others},
  title = {{AM-Bench}: A modular simulation suite and benchmark for aerial manipulation policy learning},
  booktitle = {Conference on Robot Learning (CoRL)},
  year = {2026},
  note = {Accepted; proceedings forthcoming},
}

@article{weng2024interactive,
  author = {Weng, Zehang and Zhou, Peng and Yin, Hang and Kravberg, Alexander and Varava, Anastasiia and Navarro-Alarcon, David and Kragic, Danica},
  title = {Interactive perception for deformable object manipulation},
  journal = {IEEE Robotics and Automation Letters},
  volume = {9},
  number = {9},
  pages = {7763--7770},
  year = {2024},
}

@inproceedings{xiao2026enpire,
  author = {Xiao, Wenli and Xie, Jia and Zhang, Tonghe and Lin, Haotian and Fu, Letian and Xue, Haoru and Lu, Jalen and Yang, Yi and Dai, Cunxi and Wang, Zi and others},
  title = {{ENPIRE}: Agentic robot policy self-improvement in the real world},
  booktitle = {Conference on Robot Learning (CoRL)},
  year = {2026},
  note = {Accepted; proceedings forthcoming},
}

@inproceedings{xiong2025vision,
  author = {Xiong, Haoyu and Xu, Xiaomeng and Wu, Jimmy and Hou, Yifan and Bohg, Jeannette and Song, Shuran},
  title = {Vision in action: Learning active perception from human demonstrations},
  booktitle = {Conference on Robot Learning (CoRL)},
  series = {Proceedings of Machine Learning Research},
  volume = {305},
  pages = {5450--5463},
  year = {2025},
}

@article{yadav2025modular,
  author = {Yadav, Rishabh Dev and Dantu, Swati and Pan, Wei and Sun, Sihao and Roy, Spandan and Baldi, Simone},
  title = {Modular adaptive aerial manipulation under unknown dynamic coupling forces},
  journal = {IEEE/ASME Transactions on Mechatronics},
  volume = {30},
  number = {4},
  pages = {2688--2698},
  year = {2025},
}

@inproceedings{zhao2023chat,
  author = {Zhao, Xufeng and Li, Mengdi and Weber, Cornelius and Hafez, Muhammad Burhan and Wermter, Stefan},
  title = {Chat with the environment: Interactive multimodal perception using large language models},
  booktitle = {IEEE/RSJ International Conference on Intelligent Robots and Systems (IROS)},
  pages = {3590--3596},
  year = {2023},
}

@article{zhao2023dragon,
  author = {Zhao, Moju and Okada, Kei and Inaba, Masayuki},
  title = {Versatile articulated aerial robot {DRAGON}: Aerial manipulation and grasping by vectorable thrust control},
  journal = {The International Journal of Robotics Research},
  volume = {42},
  number = {4--5},
  pages = {214--248},
  year = {2023},
}
